\pdfoutput=1

\documentclass[11pt]{article}

\usepackage[final]{acl}

\usepackage{times}
\usepackage{latexsym}
\usepackage{cleveref}
\usepackage{adjustbox}
\usepackage{listings}
\usepackage{todonotes}
\usepackage{xcolor,colortbl}

\newcolumntype{a}{>{\columncolor{blue!15}}c}

\newcommand*{\mybox}[1]{\framebox{#1}}

\definecolor{ForestGreen}{RGB}{34,139,34}

\definecolor{codegreen}{rgb}{0,0.6,0}
\definecolor{codegray}{rgb}{0.5,0.5,0.5}
\definecolor{codepurple}{rgb}{0.58,0,0.82}
\definecolor{backcolour}{rgb}{0.95,0.95,0.92}

\lstdefinestyle{mystyle}{
    backgroundcolor=\color{backcolour},   
    commentstyle=\color{codegreen},
    keywordstyle=\color{magenta},
    numberstyle=\tiny\color{codegray},
    stringstyle=\color{codepurple},
    basicstyle=\ttfamily\footnotesize,
    breakatwhitespace=false,         
    breaklines=true,                 
    captionpos=b,                    
    keepspaces=true,                 
    numbers=left,                    
    numbersep=5pt,                  
    showspaces=false,                
    showstringspaces=false,
    showtabs=false,                  
    tabsize=2
}

\usepackage[T1]{fontenc}

\usepackage[utf8]{inputenc}

\usepackage{microtype}

\usepackage{inconsolata}

\usepackage{graphicx}
\usepackage{multirow}
\usepackage{float}
\usepackage{booktabs}

\title{Watermarking Needs Input Repetition Masking}

\author{David Khachaturov, Robert Mullins\\University of Cambridge\\\And
  Ilia Shumailov, Sumanth Dathathri\\Google DeepMind\\}

\begin{document}
\maketitle
\begin{abstract}

Recent advancements in Large Language Models (LLMs) raised concerns over potential misuse, such as for spreading misinformation. 
In response two counter measures emerged: machine learning-based detectors that predict if text is synthetic, and LLM watermarking, which subtly marks generated text for identification and attribution.
Meanwhile, humans are known to adjust language to their conversational partners both syntactically and lexically. 
By implication, it is possible that humans or unwatermarked LLMs could unintentionally mimic properties of LLM generated text, making counter measures unreliable.
In this work we investigate the extent to which such conversational adaptation happens. 
We call the concept \textit{mimicry} and demonstrate that both humans and LLMs end up mimicking, including the watermarking signal even in seemingly improbable settings.
This challenges current academic assumptions and suggests that for long-term watermarking to be reliable, the likelihood of false positives needs to be significantly lower, while longer word sequences should be used for seeding watermarking mechanisms.
\end{abstract}

\section{Introduction}

\begin{figure*}[ht]
    \centering
    \includegraphics[width=0.9\linewidth]{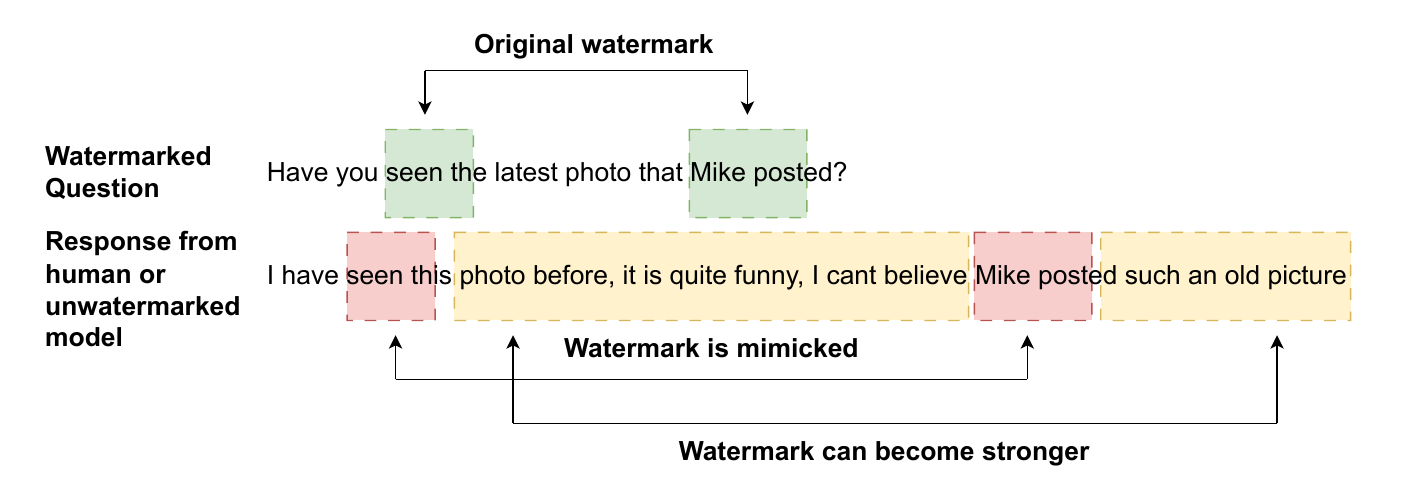}
    \caption{An intuitive description of \textit{watermark mimicry}. Here, a watermarked prompt is used together with an unwatermarked model. During a conversation parts of the original watermerk ({\color{ForestGreen}\mybox{green}}) prompt are reused by the model, leading to watermark mimicry ({\color{red}\mybox{red}}), resulting in an unwatermarked model outputting watermarked response. Importantly, the watermark can even be stronger in the response, since it can by a coincidence produce a watermark in unaffected by mimicry areas ({\color{Dandelion}\mybox{yellow}}). }
    \label{fig:mimicry_intuition}
\end{figure*}

Recent progress in Large Language Models (LLMs) enabled a large number of  applications~\citep{brown2020language,geminiteam2024gemini}, but also raised concerns about their potential misuse, particularly for spreading misinformation~\citep{openaistateactors} and impacting future machine learning (ML) models~\citep{shumailov2023curse,wyllie2024fairness}. 
To address this, ML-based detection for artificially produced text~\citep{mitchell2023detectgpt,hans2024spotting} and LLM watermarking have emerged as a promising techniques~\citep{aaronson2022my,dathathri2024deepmindwatermarks}.  
ML-based detectors work by estimating the likelihood that given text was produced by a human. Watermarks enable more precise detection~\citep{kirchenbauer2023awatermark,aaronson2023watermarking} and even attribution of the source~\citep{yoo2024advancing} by subtly embedding a unique signature within generated content.

Current LLM watermarking schemes rely on manipulating the model's internal sampling process to favour specific sequences of tokens in certain contexts~\citep{aaronson2022my,kirchenbauer2023awatermark,christ2023undetectable}. 
This allows for watermark detection by analysing the likelihood of a particular token appearing in a specific context, given a secret key. 
These methods achieve high watermark detection rates with controlable distortion to the generated text, leveraging the inherent statistical properties of language itself.

However, communication is a dynamic process. Humans, for example, adapt their language use – both in terms of vocabulary and sentence structure –  to match their conversation partner~\citep{chang2012language,mol2012adaptation}. 
This phenomenon naturally leads to the question: \textit{can the same adaptation occur during interactions with LLMs and influence watermarks and ML detectors}?  
While LLM watermarking leverages lexical and syntactic features, this very reliance creates a potential vulnerability. 
Namely, a non-watermarked conversational partner might begin to unconsciously mimic artificial language and the watermarking patterns over time, even though the partner is not watermarked. 

In this work, we introduce the concept of \textit{mimicry} and investigate the extent to which LLMs exhibit this behaviour. 
We demonstrate that through extended conversations, that LLMs exhibit \textit{mimicry}, even for the watermarks in settings where the detection probability is low according to current academic watermarking configurations. We depict the intuition in~\Cref{fig:mimicry_intuition}. 
This finding challenges the long-term effectiveness of current LLM watermarking configurations and suggests the need for exploring alternatives that take \textit{mimicry} into account and provide lower false positive rates. Fundamentally, \textit{watermark mimicry}~undermines the purpose of watermarking, since unwatermaked models are not supposed to produce watermarks. 
\section{Related Work}

\noindent \textbf{LLM Watermarking} Model watermarking emerged as a potential first party solution for attribution of synthetically generated text to its origins. Such algorithms are broadly split into distortionary~\citep{kirchenbauer2023awatermark} and non-distortianary~\citep{aaronson2023watermarking,christ2023undetectable,dathathri2024scalable}, depending on whether they change the underlying model performance. Major tech companies reportedly now watermark their models~\citep{aaronson2022my,dathathri2024deepmindwatermarks}. Importantly, current literature notes that learning from watermarked text can lead to production of watermarks~\citep{gu2024learnability,sander2024watermarking}. In this work, we show that even in-context interaction with LLMs leads to language adjustment and mimicry of the LLM watermarks.

\noindent \textbf{Third-party detectors} Alternative third party methods for synthetic data detection have also emerged. These specifically develop a machine learning model to estimate the probability that a given text sample was produced by a human~\citep{hans2024spotting,mitchell2023detectgpt}. Binoculars from~\citet{hans2024spotting} is the current state of the art for detection of artificial content, which leverages a pair of differently tuned Falcon-7B~\citep{penedo2023refinedweb} models for its calibrated prediction. 

\noindent \textbf{Conversational Adaptation} Linguistic adaptation refers to a phenomena where individuals adjust their language in response to others during a conversation~\citep{chang2012language,mol2012adaptation}. Such adaptation is extensive and covers phonology, repetition of words, syntax, and even gesturing. 

\section{Methodology}

In this paper we demonstrate that linguistic adaptation causes both humans and models to \textit{mimic watermarks} and artificiality of generated text. To show the effect we run two main experiments. First, we show that \textbf{unwatermarked} model when interacting with the \textbf{watermarked} model mimics and starts disproportionately producing the watermark. For watermarking we use two most popular schemes in the current literature: \citet{kirchenbauer2023awatermark} and~\citet{aaronson2022my}, both available through ThreeBricks~\citep{fernandez2023three}. Second, we show that a similar effect takes place in conversations with humans. Here we focus on estimating the probability that text is produced by a machine using the state-of-the-art synthetic data detector. Finally, we run experiments between the smaller Guanco-13b model and GPT-4o to observe the behavior of a more capable model. We evaluate two large publicly available datasets of multiturn human--LLM conversations. Experiments are ran on 4 $\times$ Quadro RTX 8000 and took approximately a month.

\section{Evaluation}

\begin{table*}[h]
    \centering
    \adjustbox{max width=\linewidth}{
    \begin{tabular}{llrrarrrar}
    \toprule
        \multicolumn{2}{l}{\textbf{Unwatermarked}} & 
        \multicolumn{4}{l}{\citet{aaronson2023watermarking}} & 
        \multicolumn{4}{r}{\citet{kirchenbauer2023awatermark}} \\ 
        \cmidrule(lr){3-6} \cmidrule(lr){7-10} 
        \textbf{Guanco} & \textbf{\#R} & 
        \textbf{P}rompt $\uparrow$ & \textbf{R}esponse $\downarrow$ & \textbf{P}\&\textbf{R} & \textbf{P}<\textbf{R} & 
        \textbf{P}rompt $\uparrow$ & \textbf{R}esponse $\downarrow$ & \textbf{P}\&\textbf{R} & \textbf{P}<\textbf{R} \\
    \midrule
\multicolumn{7}{l}{\color{gray}\textit{Baselines: Unwatermarked Prompt, Unwatermarked Response}}\\
\arrayrulecolor{black!30}\midrule
Theoretical & 1 & $1\%$ & $1\%$ & $0.01\%$ & & $1\%$ & $1\%$ & $0.01\%$ \\   \arrayrulecolor{black!30}\midrule
\multirow{3}{*}{7b v 7b} 
& 1 & 0.8\% & 0.9\% & 0.0\% & 0.0\% & 0.2\% & 0.8\% & 0.0\% & 0.0\% \\
& 3 & 0.7\% & 0.6\% & 0.0\% & 0.0\% & 1.6\% & 0.9\% & 0.4\% & 0.2\% \\
& 5 & 1.2\% & 0.5\% & 0.0\% & 0.0\% & 2.2\% & 1.7\% & 0.3\% & 0.3\% \\\midrule
\multirow{3}{*}{13b v 13b} 
& 1 & 1.0\% & 0.6\% & 0.2\% & 0.0\% & 0.4\% & 0.6\% & 0.1\% & 0.0\% \\
& 3 & 1.7\% & 0.5\% & 0.3\% & 0.0\% & 0.1\% & 1.1\% & 0.0\% & 0.0\% \\
& 5 & 1.4\% & 0.5\% & 0.3\% & 0.0\% & 0.2\% & 0.4\% & 0.0\% & 0.0\% \\
\arrayrulecolor{black}\midrule
\multicolumn{7}{l}{\color{gray}\textit{Watermarked Prompt, Unwatermarked Response}}\\
\arrayrulecolor{black!30}\midrule
\multirow{3}{*}{7b (w) v 7b} 
& 1 & 49.9\% & 3.7\% & 2.9\% & 0.3\% & 82.4\% & 6.3\% & 6.0\% & 0.4\% \\
& 3 & 72.0\% & 5.3\% & 4.9\% & 0.8\% & 96.7\% & 10.0\% & 10.0\% & 0.7\% \\
& 5 & 81.1\% & 6.6\% & 6.4\% & 0.7\% & 98.9\% & 13.0\% & 13.0\% & 0.7\% \\
\arrayrulecolor{black!30}\midrule
\multirow{3}{*}{7b (w) v 13b} 
& 1 & 63.4\% & 3.4\% & 2.9\% & 1.1\% & 83.8\% & 7.6\% & 7.4\% & 0.1\% \\ 
& 3 & 80.0\% & 4.7\% & 4.5\% & 0.1\% & 97.7\% & 14.4\% & 14.1\% & 0.4\% \\ 
& 5 & 89.3\% & 6.0\% & 6.0\% & 0.4\% & 99.3\% & 18.2\% & 18.2\% & 0.4\% \\

\midrule
\multirow{3}{*}{13b (w) v 7b} 
& 1 & 33.2\% & 3.9\% & 3.1\% & 0.4\% & 74.4\% & 4.6\% & 4.1\% & 0.4\% \\ 
& 3 & 54.1\% & 6.7\% & 6.2\% & 1.1\% & 93.0\% & 7.4\% & 7.3\% & 0.7\% \\ 
& 5 & 69.0\% & 8.4\% & 7.9\% & 0.5\% & 98.0\% & 10.2\% & 10.2\% & 0.8\% \\
\midrule
\multirow{3}{*}{13b (w) v 13b} 
& 1 & 30.6\% & 3.2\% & 2.0\% & 0.3\% & 73.1\% & 5.8\% & 5.6\% & 0.4\% \\
& 3 & 53.2\% & 6.3\% & 5.2\% & 0.7\% & 90.5\% & 8.3\% & 8.3\% & 0.4\% \\
& 5 & 63.1\% & 8.1\% & 7.2\% & 0.9\% & 94.7\% & 11.2\% & 11.0\% & 0.4\% \\
\arrayrulecolor{black}\midrule
\multicolumn{7}{l}{\color{gray}\textit{Watermarked Prompt via Guanco-13b, Unwatermarked Response with GPT-4o}}\\
\arrayrulecolor{black!30}\midrule
\multirow{3}{*}{ngram=2, t=0.8} 
& 1 & 56.9\% ($\pm$ 3.9\%) & 0.8\% ($\pm$ 0.5\%) & 0.5\% ($\pm$ 0.3\%) & 7.3\% ($\pm$ 0.9\%) & 83.6\% ($\pm$ 4.4\%) & 1.3\% ($\pm$ 0.3\%) & 1.2\% ($\pm$ 0.2\%) & 5.0\% ($\pm$ 1.2\%) \\
& 3 & 88.2\% ($\pm$ 1.1\%) & 1.8\% ($\pm$ 0.8\%) & 1.6\% ($\pm$ 0.8\%) & 3.4\% ($\pm$ 0.2\%) & 97.6\% ($\pm$ 1.6\%) & 1.6\% ($\pm$ 0.6\%) & 1.6\% ($\pm$ 0.5\%) & 0.3\% ($\pm$ 0.2\%) \\
& 5 & 93.7\% ($\pm$ 0.6\%) & 2.5\% ($\pm$ 1.3\%) & 2.0\% ($\pm$ 1.1\%) & 2.7\% ($\pm$ 0.3\%) & 99.4\% ($\pm$ 0.4\%) & 2.4\% ($\pm$ 0.9\%) & 2.4\% ($\pm$ 0.9\%) & 0.1\% ($\pm$ 0.1\%) \\
\midrule
\multirow{3}{*}{ngram=4, t=0.8} 
& 1 & 39.7\% ($\pm$ 10.1\%) & 0.6\% ($\pm$ 0.2\%) & 0.1\% ($\pm$ 0.0\%) & 25.4\% ($\pm$ 6.7\%) & 79.9\% ($\pm$ 9.5\%) & 0.8\% ($\pm$ 0.5\%) & 0.6\% ($\pm$ 0.3\%) & 10.3\% ($\pm$ 6.7\%) \\
& 3 & 66.2\% ($\pm$ 15.9\%) & 0.8\% ($\pm$ 0.3\%) & 0.5\% ($\pm$ 0.3\%) & 16.3\% ($\pm$ 7.1\%) & 93.8\% ($\pm$ 6.0\%) & 1.5\% ($\pm$ 0.8\%) & 1.4\% ($\pm$ 0.7\%) & 2.5\% ($\pm$ 2.8\%) \\
& 5 & 74.1\% ($\pm$ 14.1\%) & 1.1\% ($\pm$ 0.4\%) & 0.9\% ($\pm$ 0.3\%) & 14.6\% ($\pm$ 6.5\%) & 95.9\% ($\pm$ 5.2\%) & 1.8\% ($\pm$ 0.9\%) & 1.7\% ($\pm$ 0.8\%) & 0.7\% ($\pm$ 0.7\%) \\
\midrule
\multirow{3}{*}{ngram=6, t=0.8} 
& 1 & 49.3\% ($\pm$ 13.1\%) & 1.1\% ($\pm$ 0.4\%) & 0.6\% ($\pm$ 0.2\%) & 17.5\% ($\pm$ 6.2\%) & 79.1\% ($\pm$ 8.0\%) & 1.2\% ($\pm$ 0.1\%) & 1.0\% ($\pm$ 0.2\%) & 10.7\% ($\pm$ 5.4\%) \\
& 3 & 76.2\% ($\pm$ 11.2\%) & 1.2\% ($\pm$ 0.7\%) & 0.8\% ($\pm$ 0.3\%) & 7.3\% ($\pm$ 4.3\%) & 93.9\% ($\pm$ 4.3\%) & 0.8\% ($\pm$ 0.3\%) & 0.8\% ($\pm$ 0.2\%) & 2.5\% ($\pm$ 2.2\%) \\
& 5 & 84.2\% ($\pm$ 8.3\%) & 1.7\% ($\pm$ 0.7\%) & 1.5\% ($\pm$ 0.4\%) & 5.4\% ($\pm$ 3.3\%) & 96.9\% ($\pm$ 2.9\%) & 0.9\% ($\pm$ 0.4\%) & 0.9\% ($\pm$ 0.4\%) & 1.4\% ($\pm$ 1.6\%) \\
\midrule
\multirow{3}{*}{ngram=8, t=0.8} 
& 1 & 44.3\% ($\pm$ 18.1\%) & 1.3\% ($\pm$ 0.4\%) & 0.7\% ($\pm$ 0.5\%) & 22.9\% ($\pm$ 16.7\%) & 77.4\% ($\pm$ 11.7\%) & 0.9\% ($\pm$ 0.2\%) & 0.7\% ($\pm$ 0.1\%) & 14.5\% ($\pm$ 8.3\%) \\
& 3 & 67.7\% ($\pm$ 21.8\%) & 1.6\% ($\pm$ 0.8\%) & 1.2\% ($\pm$ 0.9\%) & 10.3\% ($\pm$ 9.1\%) & 91.1\% ($\pm$ 6.2\%) & 1.3\% ($\pm$ 0.2\%) & 1.2\% ($\pm$ 0.2\%) & 3.7\% ($\pm$ 2.6\%) \\
& 5 & 78.0\% ($\pm$ 18.2\%) & 1.6\% ($\pm$ 1.0\%) & 1.4\% ($\pm$ 1.0\%) & 9.6\% ($\pm$ 10.0\%) & 95.0\% ($\pm$ 3.9\%) & 1.0\% ($\pm$ 0.2\%) & 1.0\% ($\pm$ 0.2\%) & 1.4\% ($\pm$ 1.2\%) \\
\arrayrulecolor{black}\bottomrule
    \end{tabular}}
    \caption{Table shows proportion of text that has a watermark. Here a watermarked LLM model (Temperature=1, watermark ngram=4, threshold 0.01, averaged over three random keys (large variance)) interacts with a non-watermarked LLM model (Temperature=0.8). \textbf{P}rompt refers to the \textit{watermarked prompt} from a first model. \textbf{R}esponse refers to the \textit{response of an unwatermarked model} to a watermarked prompt. \textbf{\#R} refers to a number of responses concatenated together. \textbf{P\&R} reports a proportion of cases where watermarked prompt -- i.e. watermark score < 0.01 -- results in a watermarked response -- watermark score < 0.01. \textbf{P<R} reports the proportion of cases where watermark of the response from an unwatermarked model is stronger than the watermark in the prompt. Original conversation prompt is listed in~\Cref{sec:original_prompt}. Expanded table in presented in~\Cref{tab:combined_wms_expanded} in Appendix. }
    \label{tab:combined_wms}
\end{table*}

\subsection{LLMs conversing with LLMs}\label{sec:llm-llm-convo}

In this subsection we focus on the conversation between a pair of models -- one watermarked with either~\citet{kirchenbauer2023awatermark}~or~\citet{aaronson2023watermarking} schemes; while the second model uses no watermark. We utilise Guanco-7B, Guanco-13B~\citep{dettmers2023qlora} models and run the watermarking model in three temperature configurations T=[0, 0.5, 1]. We show examples of conversations in~\Cref{sec:conversation_examples} and the prompt in~\Cref{sec:original_prompt}. 

\noindent \Cref{tab:combined_wms} shows the effects of \textit{watermark mimicry}. The flow of conversations are shown in~\Cref{fig:aaronson_7b,fig:aaronson_13b} for~\citet{aaronson2023watermarking} and in~\Cref{fig:kirch_7b,fig:kirch_13b} for~\citet{kirchenbauer2023awatermark}. We find that for both schemes mimicry appears, albeit the effect is stronger for~\citeauthor{kirchenbauer2023awatermark}. For example, with~\citeauthor{aaronson2022my} scheme Guanco 13b unwatermarked model outputs watermarked response in 4\% of cases when used with temperature 1, when computed over 5 consecutive responses. While in the same setting, \citeauthor{kirchenbauer2023awatermark} scheme outputs a watermarked response 12.9\% of time. In \Cref{fig:kirch_13b_deduped} we confirm that when input repetition masking is applied text is no longer watermarked. 

\noindent \Cref{tab:combined_wms} also presents the results of conversation between the smaller Guanco-13b model with fixed temperature $t=0.8$ and varying ngrams $2,4,6,8$ and GPT-4o, averaged over 3 different hash keys. We can see that the mimicry observed for smaller models does not hold.

\noindent We hypothesize that the absence of mimicry in larger models is attributable to their superior language modeling capabilities. Larger models, such as the evaluated GPT-4o exhibit stronger generalization and appear less susceptible to adopting local lexical or syntactic patterns present in the prompt. They may rely less on shallow repetition mechanisms and instead generate responses based on broader contextual understanding. Furthermore, models like GPT-4o may internally smooth over deterministic token-level irregularities introduced by watermarking schemes, reducing the likelihood of unintentionally reproducing embedded patterns.

\subsection{Third party detection of generated data}
\label{sec:third_party_eval}
In this section we focus on a setting where a human has a conversation with an LLM. Here we make no assumptions about existence of watermarking and focus on the best publicly available third-party detector Binoculars~\citep{hans2024spotting}. We reuse the best parameters from~\citeauthor{hans2024spotting} and calibrate for 1\% false positives rates. We use Binoculars to estimate if during the conversation with an LLMs humans adjust their language and `mimic' the LLM, i.e. human language gets identified as LLM. \\

\begin{figure*}
\centering
\includegraphics[width=0.33\textwidth]{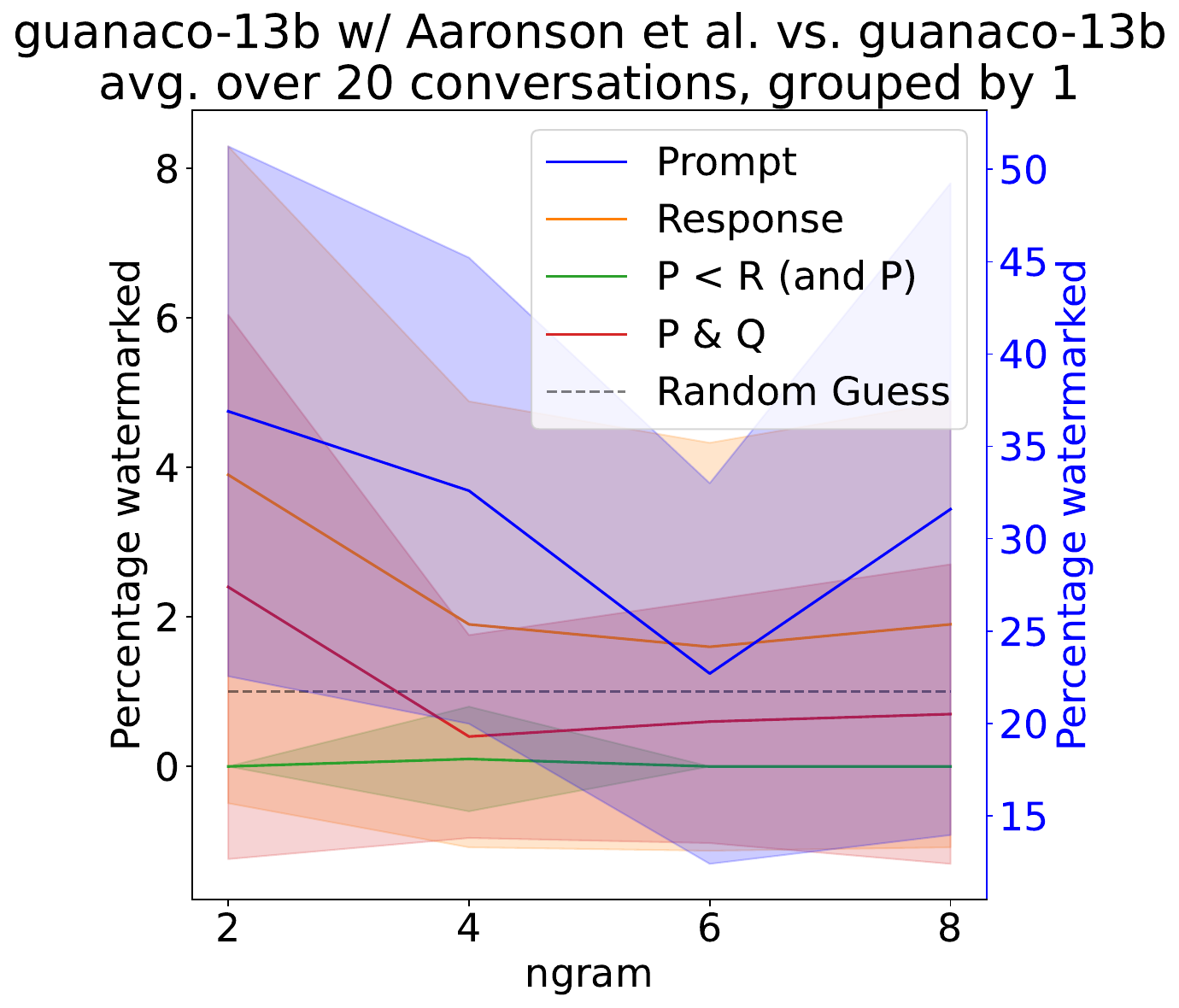}\hfill
\includegraphics[width=0.33\textwidth]{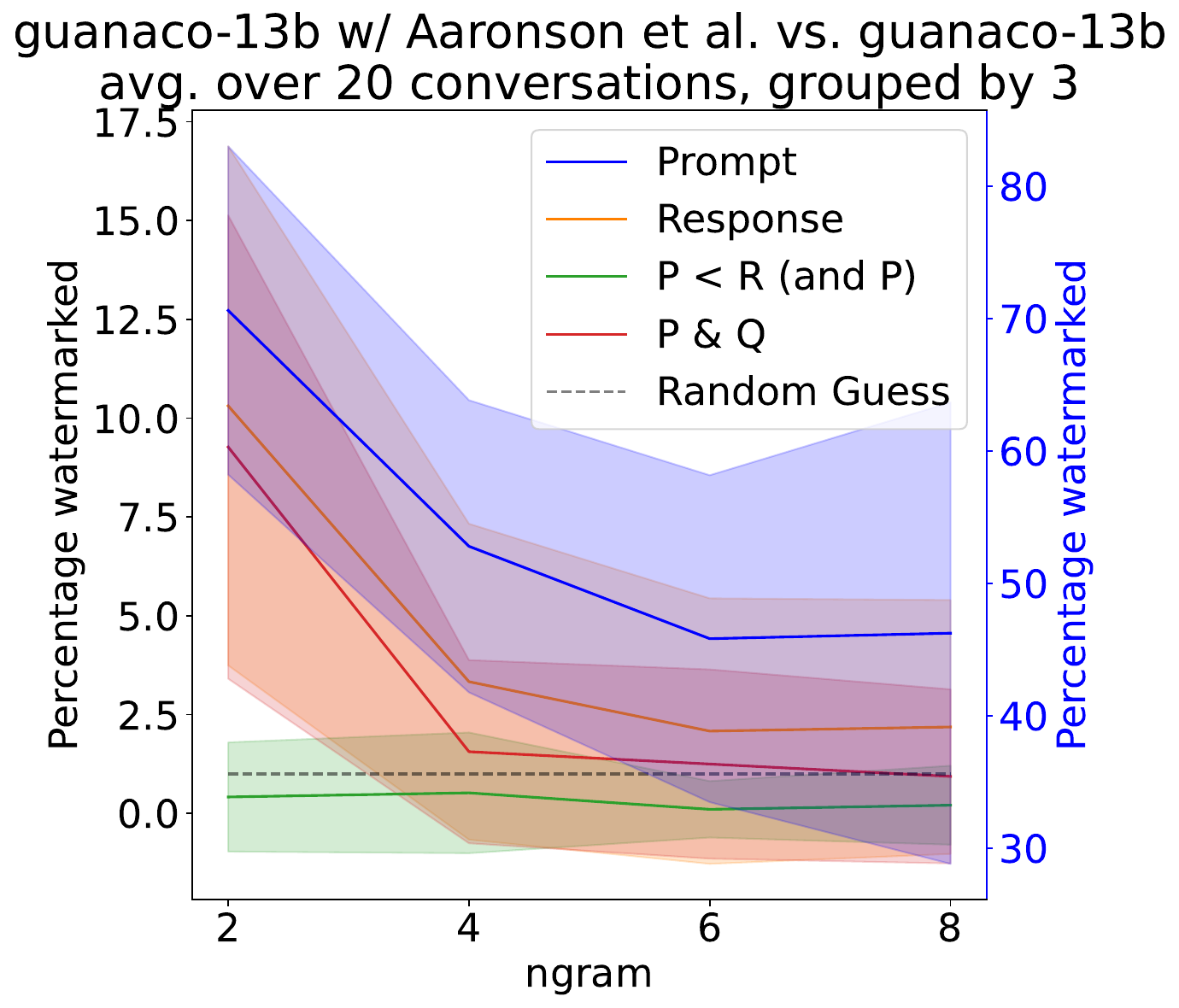}\hfill
\includegraphics[width=0.33\textwidth]{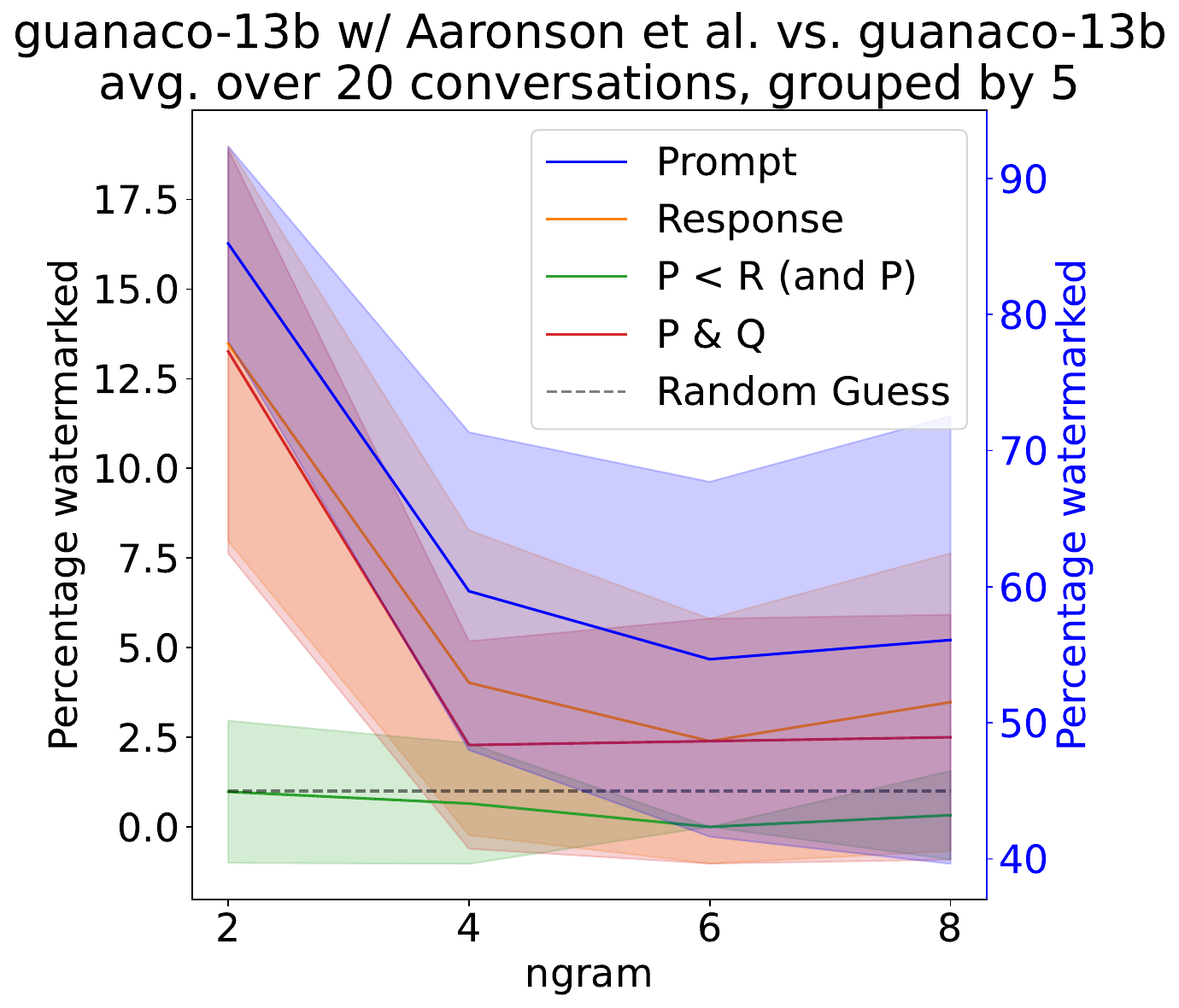}
\caption{\citet{aaronson2022my} scheme, changing ngram size, {\color{blue}blue} shows percentage of watermarked prompts, {\color{orange}orange} shows percentage of watermarked responses {\color{green}green} shows percentage of watermarked responses where response watermark is stronger than in the prompt, {\color{red}red} shows percentage of cases with both prompt and response watermarked. }
\label{fig:n-gram-impact-aaranson}
\end{figure*}

\begin{figure}
    \centering
    \includegraphics[width=0.85\linewidth]{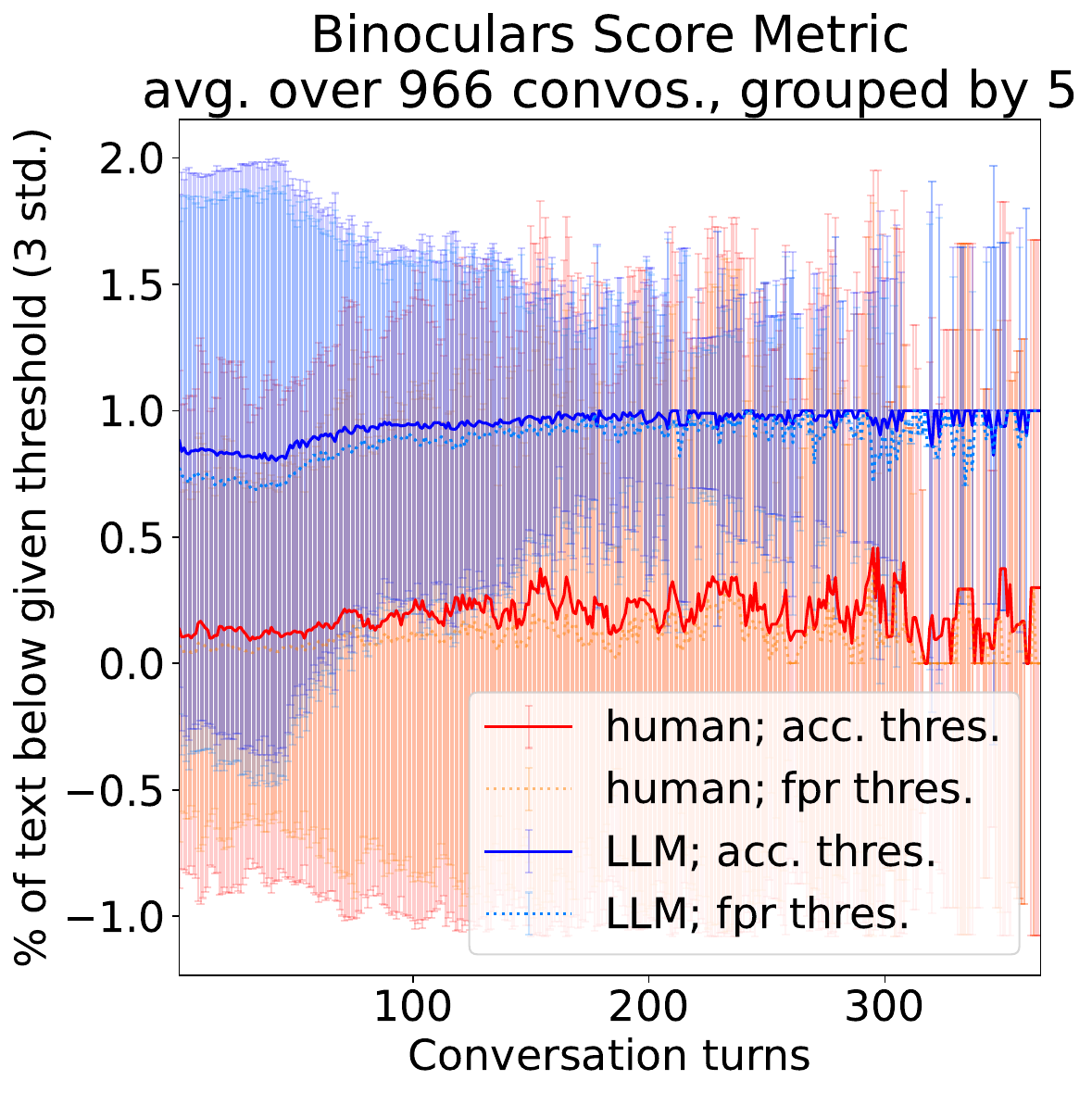}
    \caption{Human--LLM dialogues (split in \Cref{fig:sharegpt_wildchat}), filtered to contain long conversations in English. 520 are from ShareGPT, filtered for 100+ turns (100 human, 100 LLM). 446 are from WildChat dataset, filtered for 50+ turns (100 human, 100 LLM).}
    \label{fig:human-llm-convo-dataset}
\end{figure}

\noindent \textbf{LLM-Human conversations:} Here we evaluate conversations dataset available through huggingface \texttt{ar852/scraped-chatgpt-conversations}, \texttt{theblackcat102/sharegpt-english}, filtered to only keep conversations with at least 100 turns. \Cref{fig:human-llm-convo-dataset} shows the results. We find that during the conversations humans adjust their language and variance of scores increases over the dialogue. In other words, over a dialogue humans tend to produce more speech that appears LLM-like for Binoculars. The opposite effect takes place for the LLM, where variance is lost over the conversation. 



\section{Discussion}

In this paper we demonstrated that LLM-produced input influences how language is used by unwatermarked models and humans. Our results have methodological as well as practical implications. First, methodologically we show that for watermarking it is important to account for linguistic adaption when evaluating performance. Particularly so when the prompts used for evaluations were generated by other models. One of the ways to deal with mimicry is to incorporate repetition masking that is used in some watermarking mechanisms already~\citep{hu2023unbiased}, extending its use to the prompts. Second, we need to assume stronger detection thresholds and more realistic configurations for watermarking. This means adjusting false positive rates to be lower and, for example, choosing ngram lengths that result in less robust watermarks. \Cref{fig:n-gram-impact-aaranson} shows how amount of watermark mimicry for~\citet{aaronson2023watermarking} changes as a function of ngram. We show more results in Appendix~\Cref{fig:aaronson_ngrams_13_7,fig:aaronson_ngrams_7_13,fig:aaronson_ngrams_13_13,fig:aaronson_ngrams_7_7}. Here, we observe that as ngrams get longer mimicry happens less, yet it comes at a cost of watermarking robustness.

\section{Conclusion}

Our work highlights a critical challenge for LLM provenance -- \textit{mimicry}. We demonstrate that small LLMs exhibit adaptation during conversations, potentially replicating language artificiality and surprisingly even the watermark signal. This phenomenon occurs less prominently in larger models.
This undermines the effectiveness of current watermarking schemes that rely on easily detectable deviations in token probabilities.
To ensure the long-term viability of LLM watermarking, we propose further research avenues. 
First, investigating watermarking techniques that provide significantly lower false positive rates is crucial. 
Second, exploring alternative watermarking methods that leverage properties less susceptible to mimicry, such as semantic coherence or stylistic elements. 
By addressing these challenges, we can ensure that LLM watermarking remains a valuable tool for long-term mitigation of the misuse of LLMs.

\section{Limitations}
In this section we discuss limitations of the evaluation above. First, our work only considers two of the most popular schemes and it is not clear if all other watermarking mechanisms will exhibit mimicry. Intuitively we expect it to still happen, consider the following pathological example. Imagine an unwatermarked model quoting watermarked text from the input -- in this case as long as quotation happens it should always trigger the watermark detector. Yet, the example above provides no indication to the degree of unnoticeable mimicry. Second, our work only considered watermarks used with three keys and computing expectation over more keys may reduce currently large performance variance. We still nonetheless believe that mimicry will be exhibited, based on the examples and the shear magnitude of mimicry observed in our experiments with the three random keys. Third, we only considered models of size 7b and 13b, while larger models may exhibit less mimicry and generalise to text better. It can clearly be seen in the examples in~\Cref{sec:conversation_examples}, where the text is comprehensible, the quality is not on par with the best currently available models. This choice is motivated by limited resources. That said, initial results with larger models are presented in~\Cref{sec:llm-llm-convo} and indeed support our intuition. Fourth, evaluation in~\Cref{sec:third_party_eval} assumes that data available through public datasets is reliable to make adaptation judgements. Fifth, following from the previous point, it is not clear to what degree mimicry affects all human conversations, and not just specific type of prolonged discussions. If discussions are short and only a few sentences are exchanges, it is not obvious that mimicry would always happen. More thorough evaluation is required to find precise type of linguistic adaptation that enables watermark mimicry. 

\section{Ethical Considerations}
Our work addresses reliability of text watermarking mechanisms -- 
an important tool that enables trustworthy machine learning. We find that to make well informed text provenance decisions, \textit{mimicry} and linguistic adaptation should be taken into account. We highlight an inherent assumption of text watermarking and suggest future venues for addressing the discovered limitation. We view our work as an important step towards making text watermarking more reliable, while also making methodologies of watermark evaluations more transparent and fair. 

\section*{Contributions}
All experiments were conducted by D.K. at the home institution. I.S. and S.D. participates purely in advisory capacity and conducted no experiments.

\bibliography{references}

\newpage
\appendix

\section{Multi-turn Conversations}
\label{sec:appendix}

\begin{figure}[t]
\centering
\includegraphics[width=0.4\textwidth]{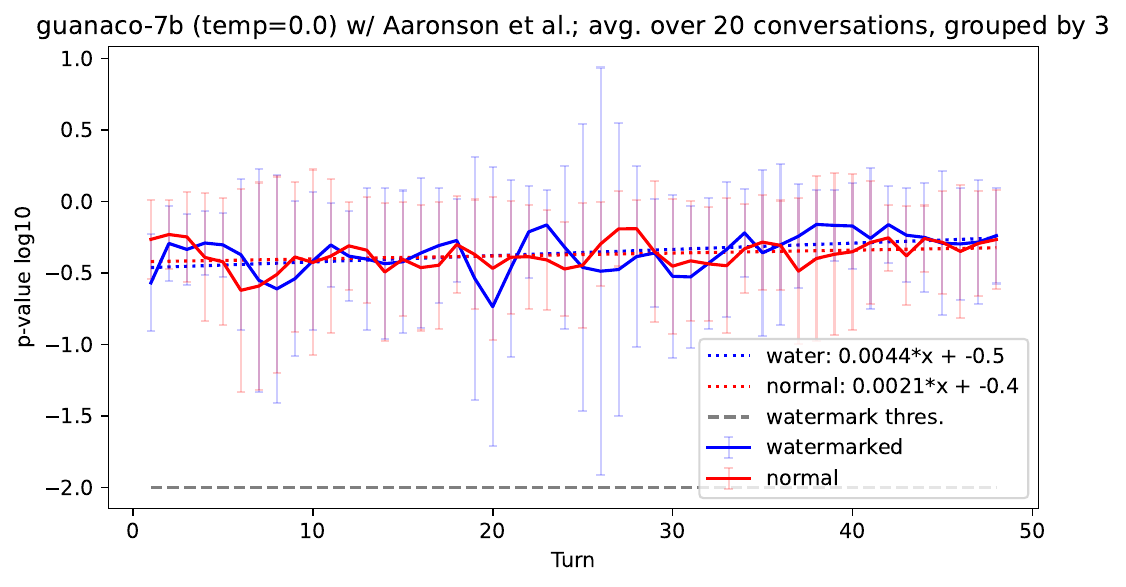}\hfill
\includegraphics[width=0.4\textwidth]{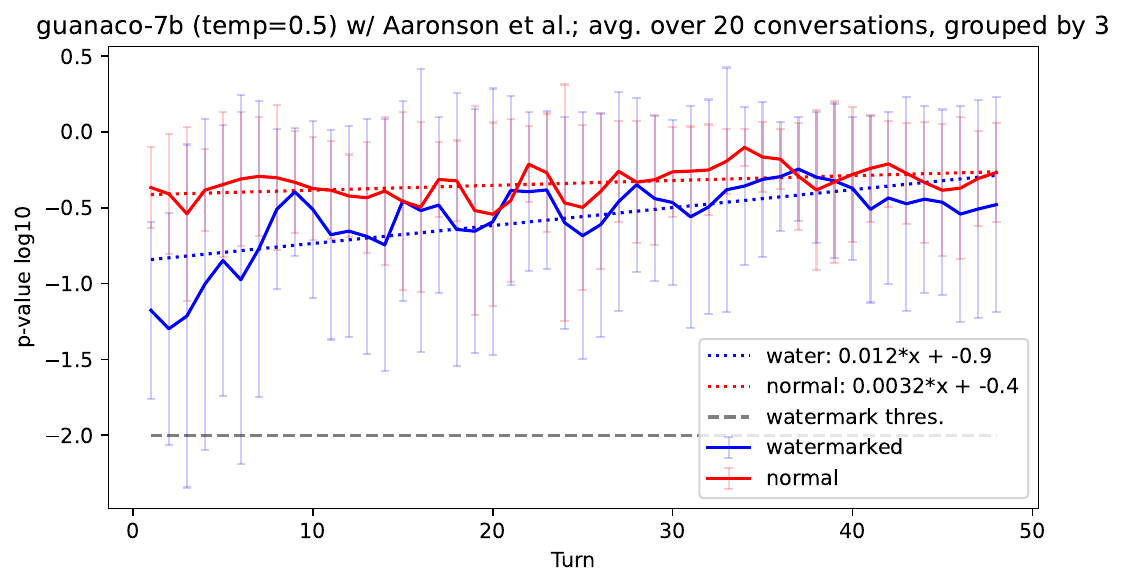}\hfill
\includegraphics[width=0.4\textwidth]{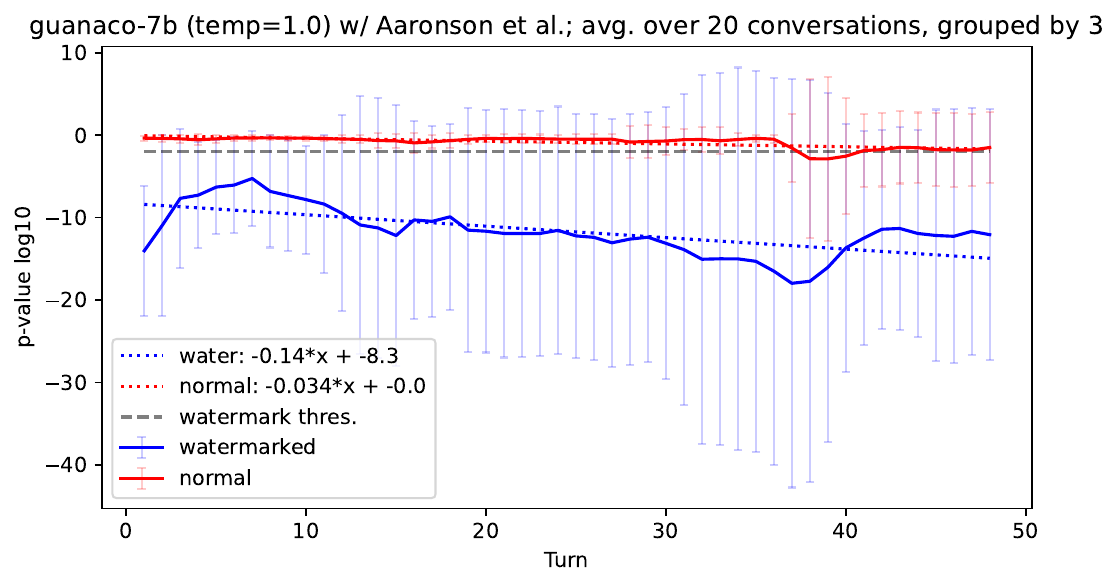}
\caption{Aaronson. watermarking with Guanco-7b and varying temperatures}
\label{fig:aaronson_7b}
\end{figure}

\begin{figure}[t]
\centering
\includegraphics[width=0.4\textwidth]{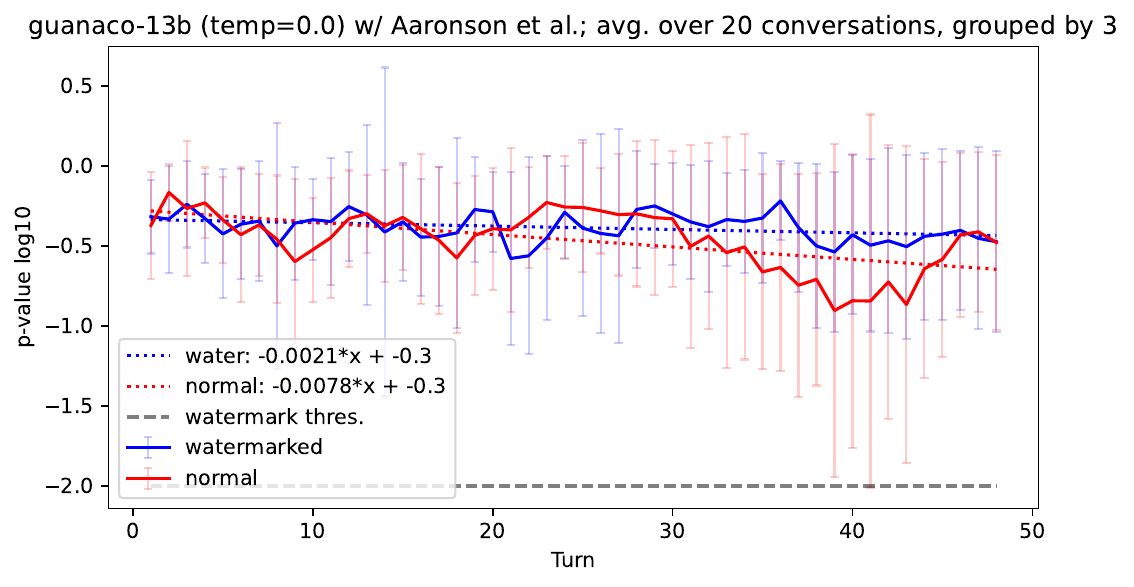}\hfill
\includegraphics[width=0.4\textwidth]{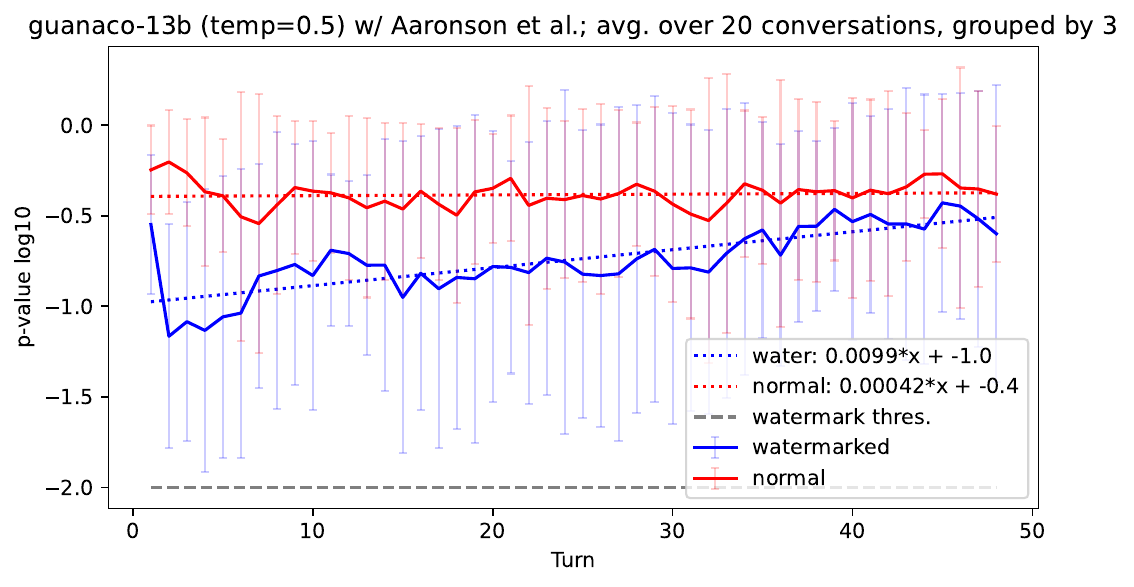}\hfill
\includegraphics[width=0.4\textwidth]{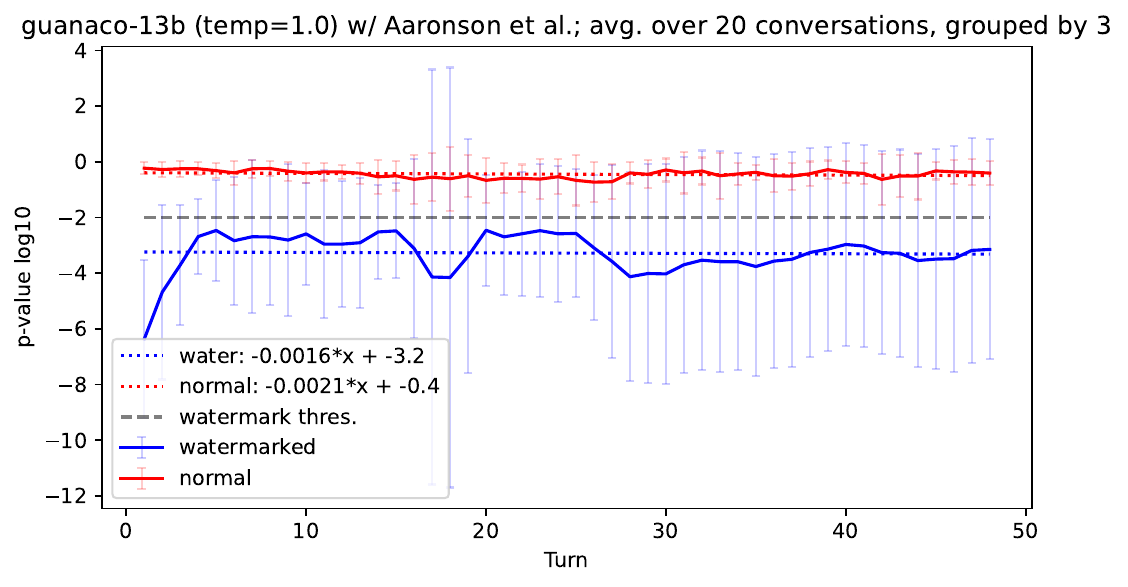}
\caption{Aaronson. watermarking with Guanco-13b and varying temperatures}
\label{fig:aaronson_13b}
\end{figure}

\begin{figure}[H]
\centering
\includegraphics[width=0.4\textwidth]{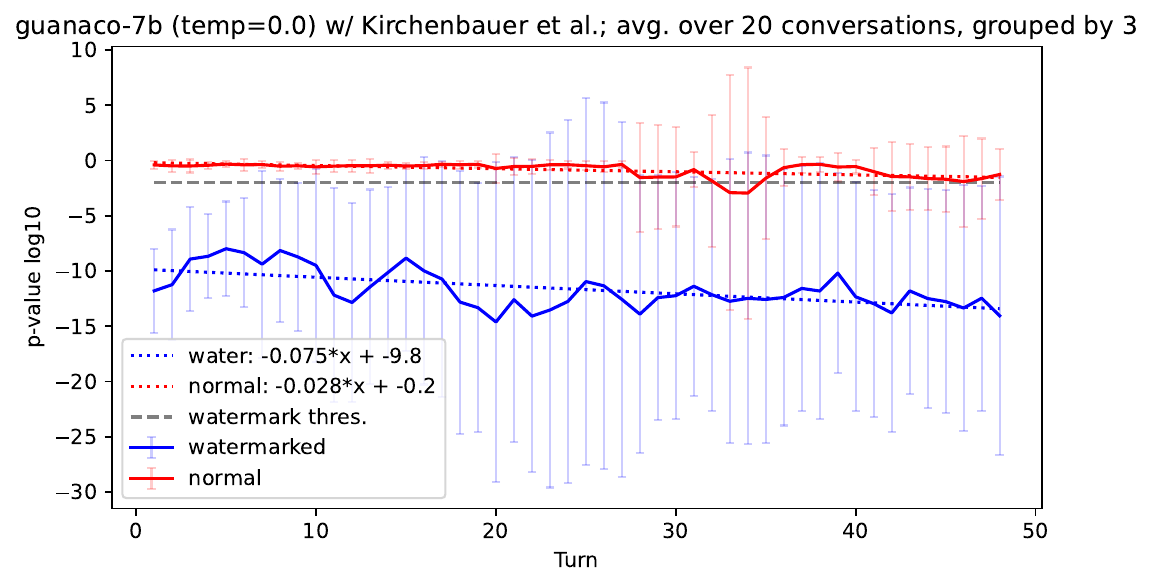}\hfill
\includegraphics[width=0.4\textwidth]{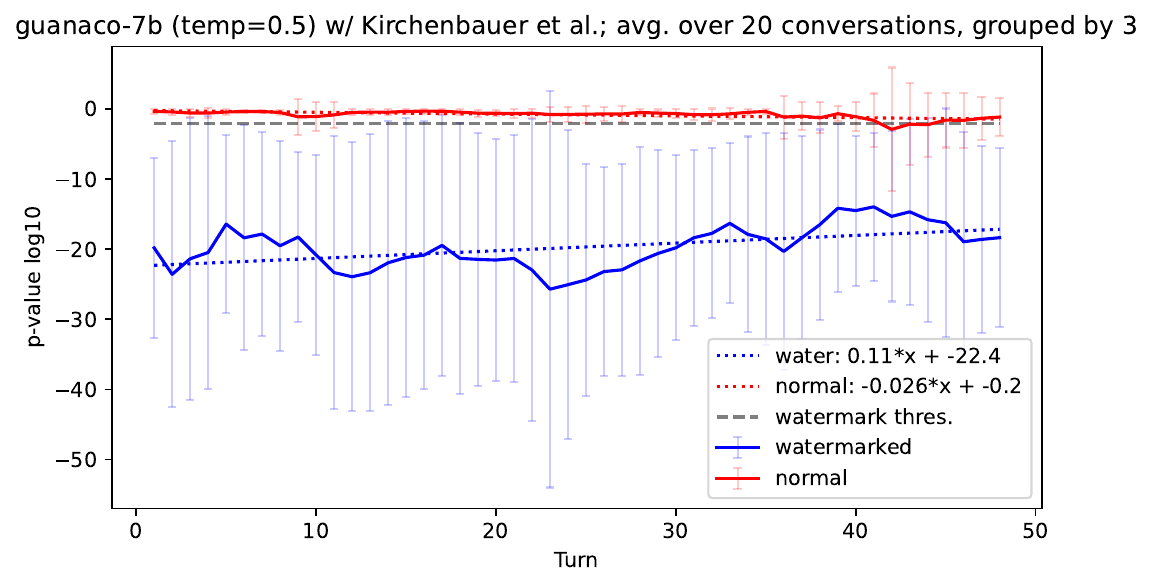}\hfill
\includegraphics[width=0.4\textwidth]{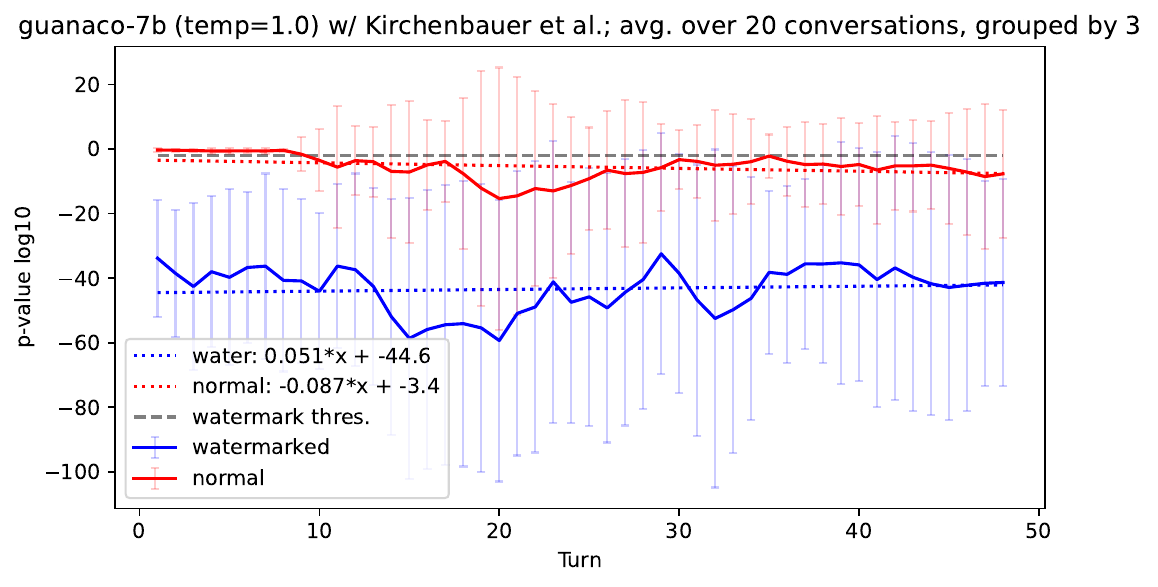}
\caption{\citet{kirchenbauer2023awatermark} watermarking with Guanco-7b and varying temperatures.}
\label{fig:kirch_7b}
\end{figure}

\begin{figure}[H]
\centering
\includegraphics[width=0.4\textwidth]{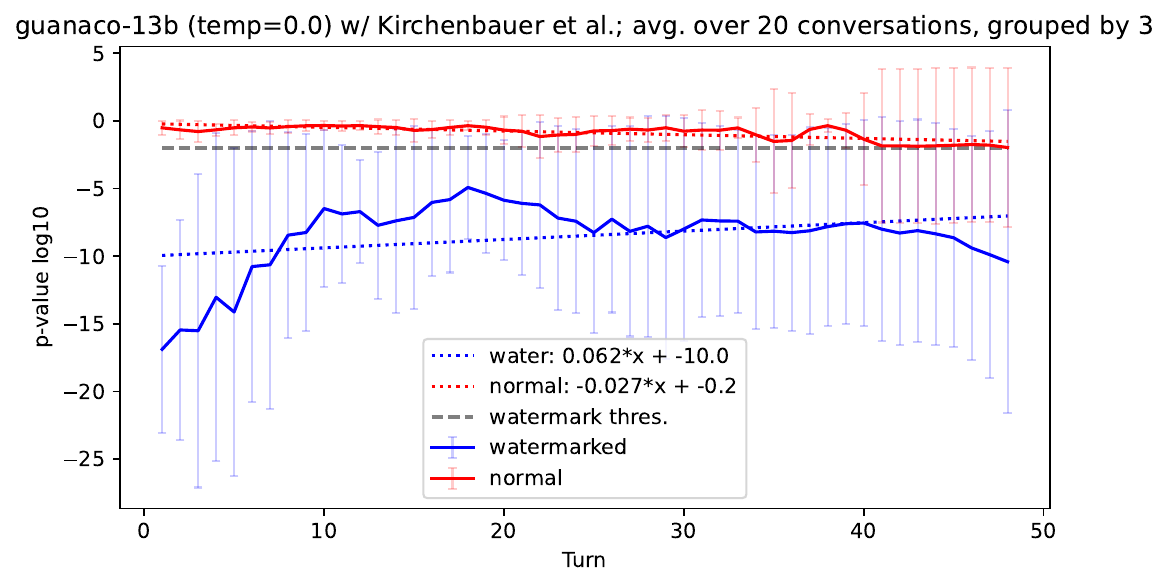}\hfill
\includegraphics[width=0.4\textwidth]{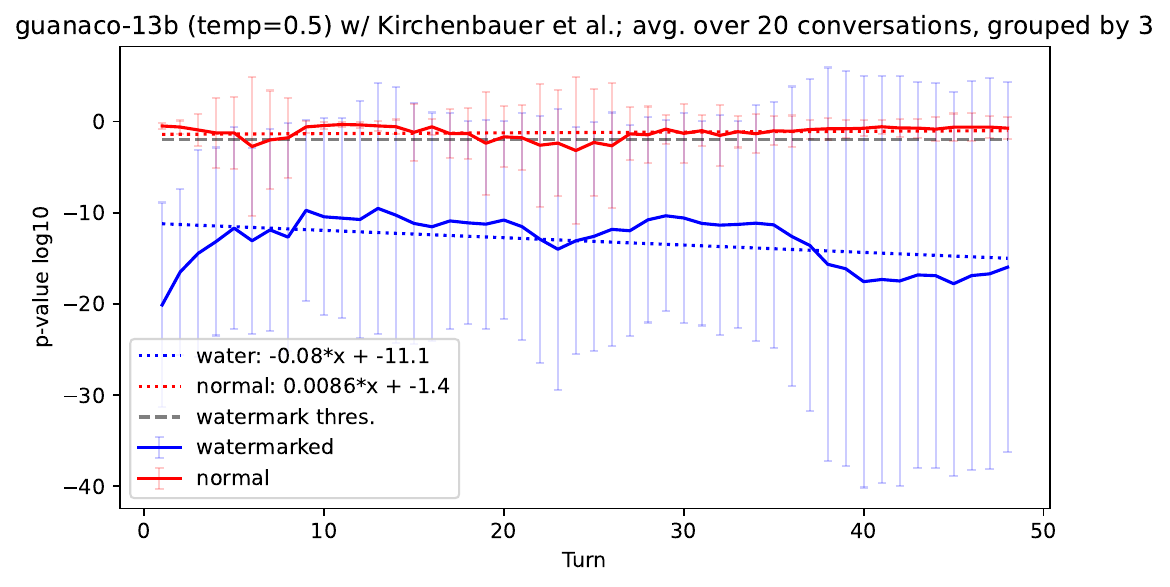}\hfill
\includegraphics[width=0.4\textwidth]{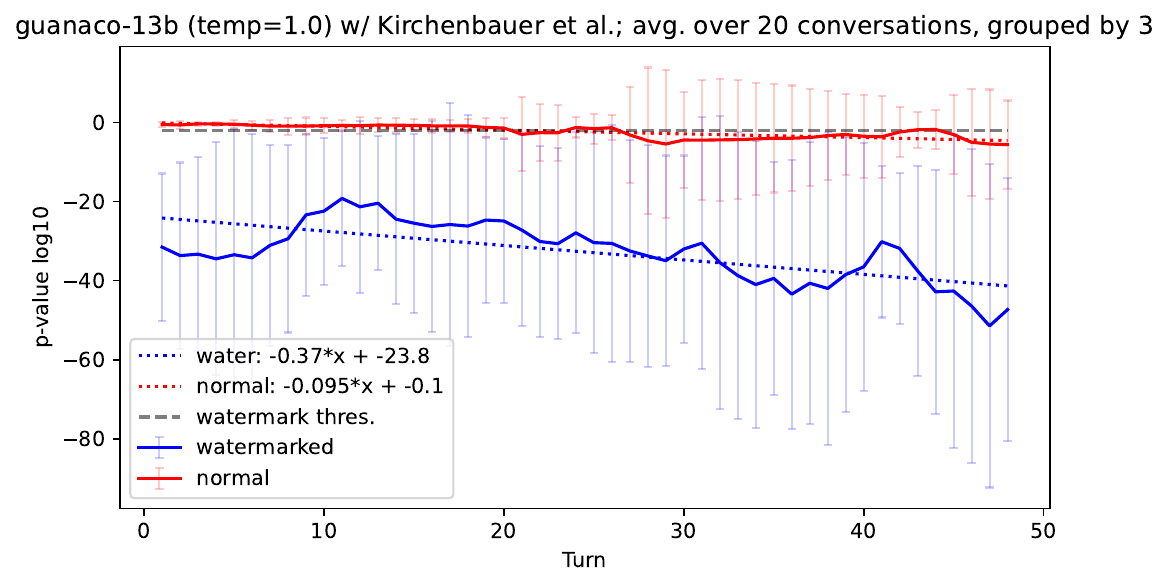}
\caption{\citet{kirchenbauer2023awatermark} watermarking with Guanco-13b and varying temperatures.}
\label{fig:kirch_13b}
\end{figure}

\begin{figure}[t]
\centering
\includegraphics[width=0.4\textwidth]{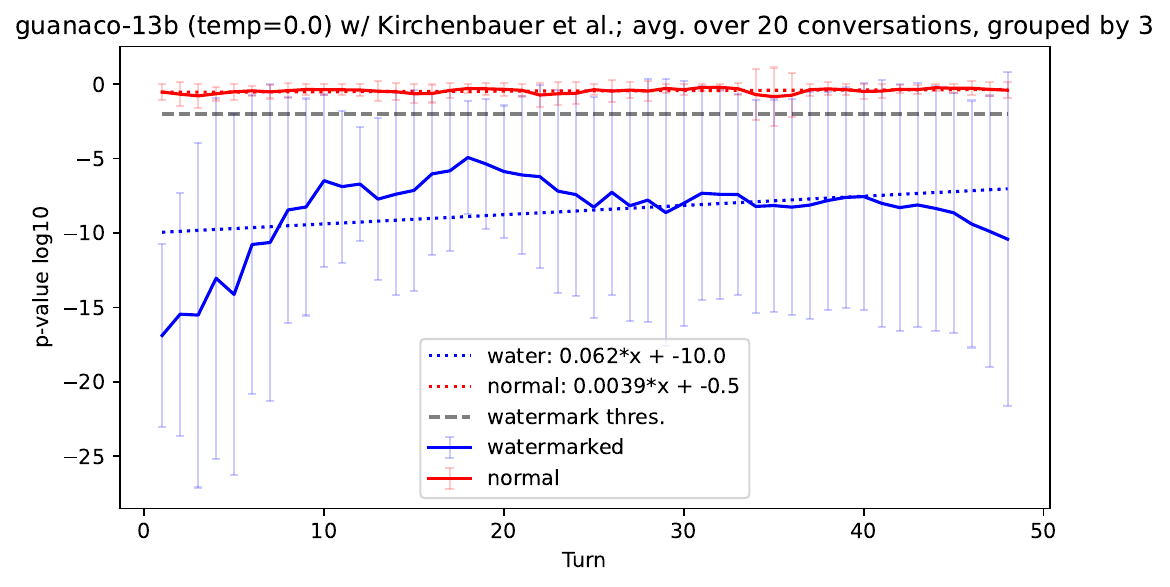}\hfill
\includegraphics[width=0.4\textwidth]{figures/13b_kirch-llm-to-llm-t0-g3_DEDUP.pdf}\hfill
\includegraphics[width=0.4\textwidth]{figures/13b_kirch-llm-to-llm-t0-g3_DEDUP.pdf}
\caption{\citet{kirchenbauer2023awatermark} watermarking with Guanco-13b and varying temperatures, de-duplicating ngrams between the watermarked input and the non-watermarked output. We note that this de-duplication removes the observed mimicry behavior.}
\label{fig:kirch_13b_deduped}
\end{figure}

\begin{figure*}
\centering
\includegraphics[width=0.3\textwidth]{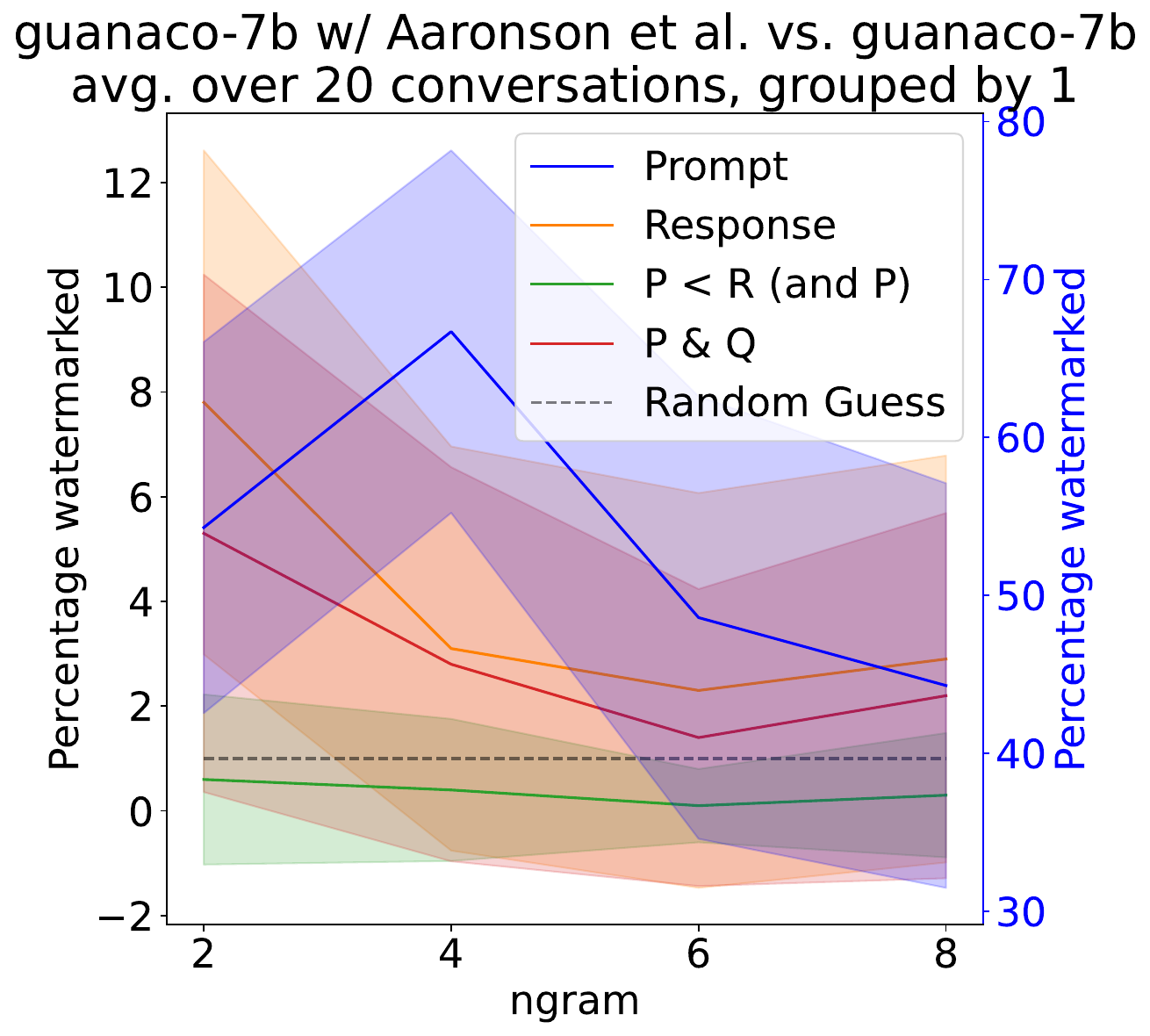}\hfill
\includegraphics[width=0.3\textwidth]{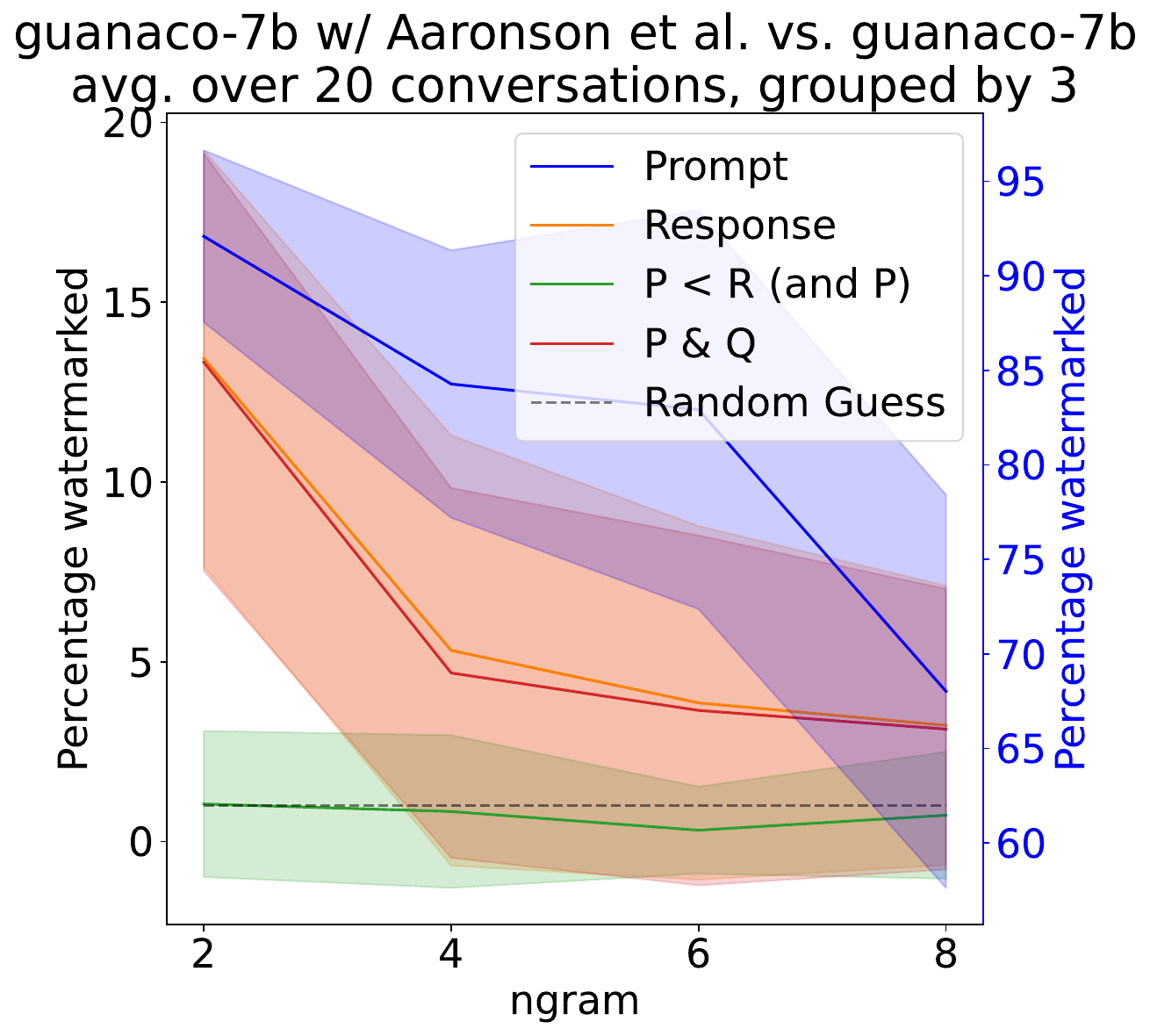}\hfill
\includegraphics[width=0.3\textwidth]{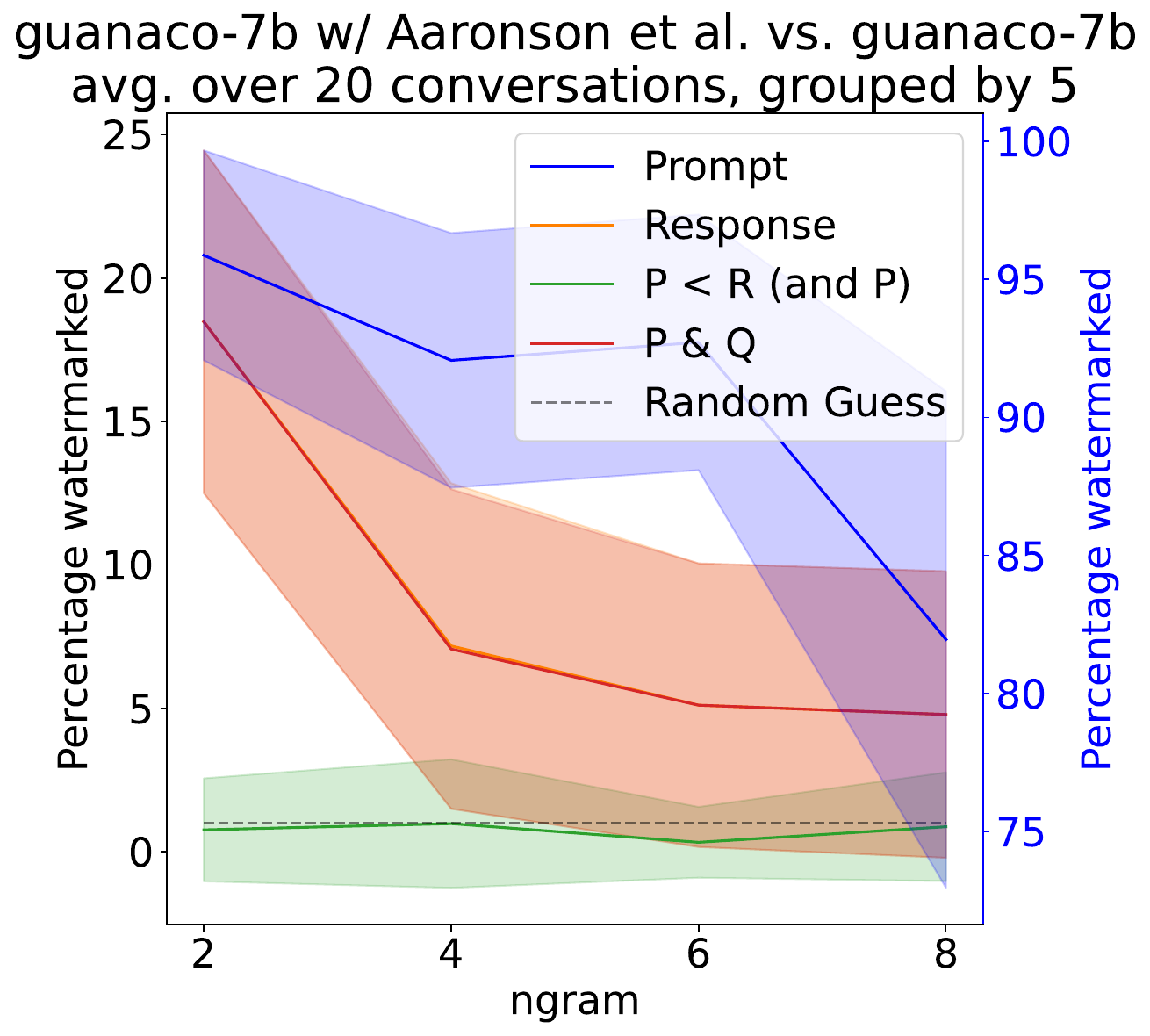}
\caption{\citet{aaronson2022my} with Guanco-7b (w) v 7b and varying ngrams, T=1}
\label{fig:aaronson_ngrams_7_7}
\end{figure*}

\begin{figure*}
\centering
\includegraphics[width=0.3\textwidth]{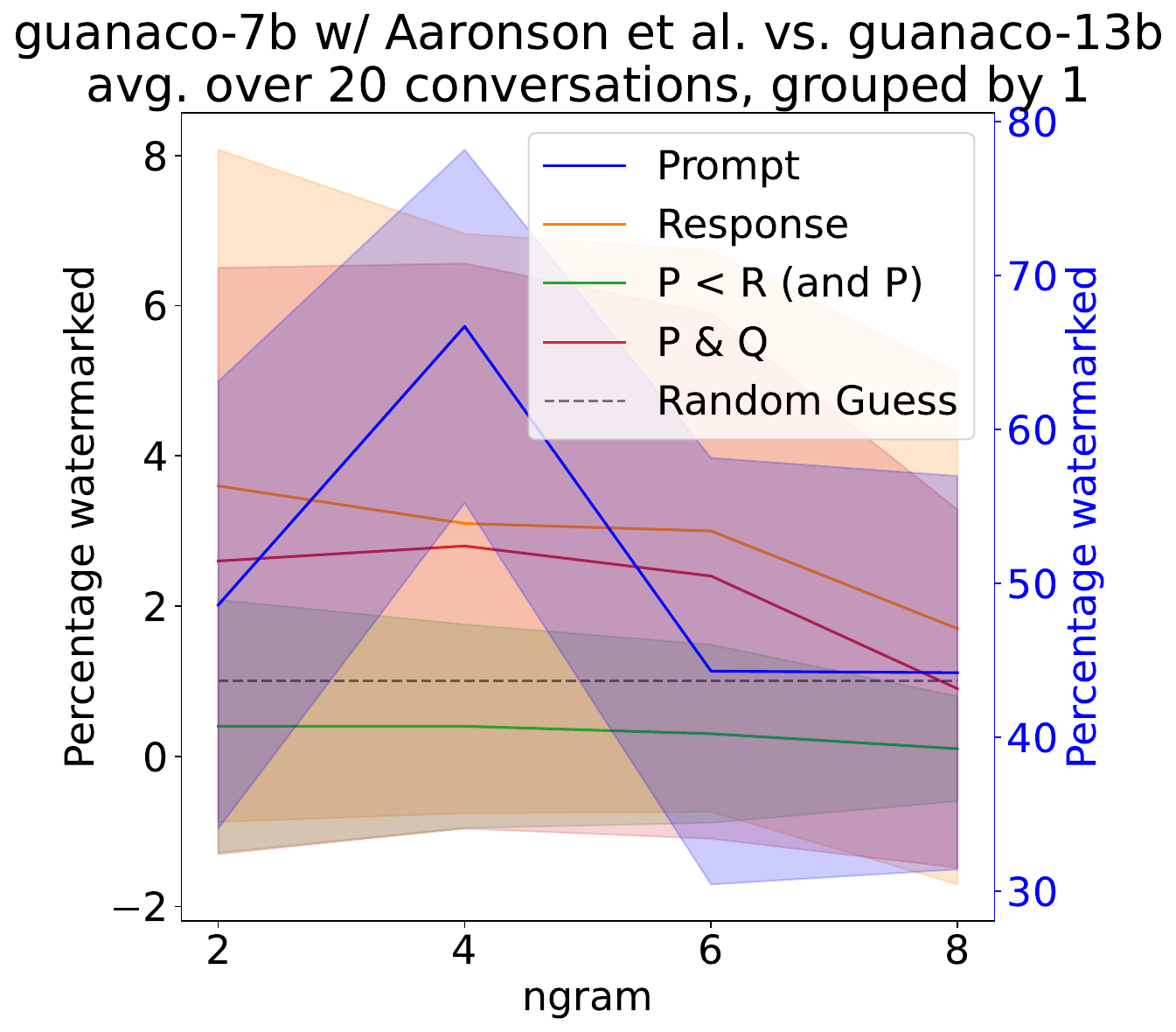}\hfill
\includegraphics[width=0.3\textwidth]{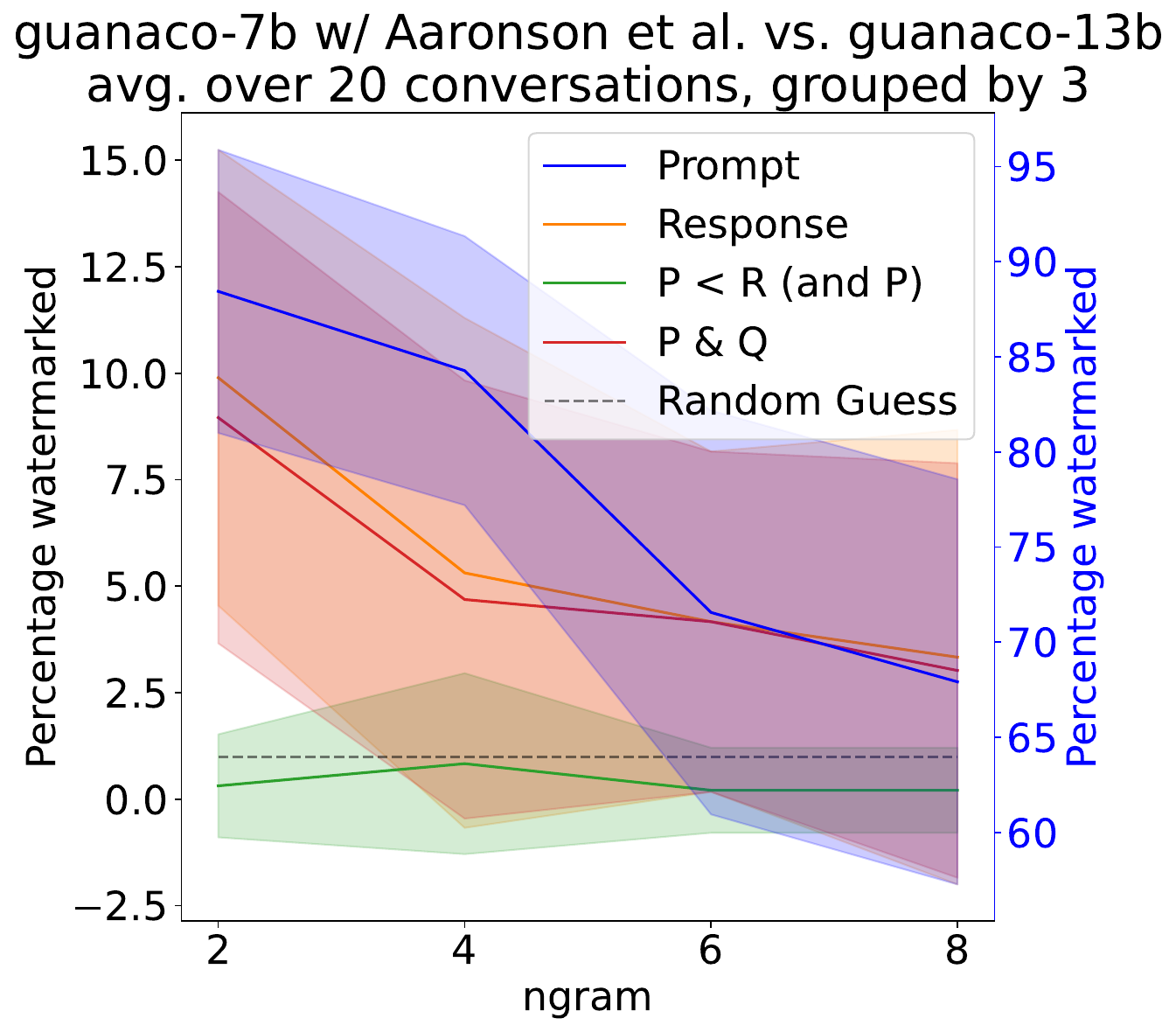}\hfill
\includegraphics[width=0.3\textwidth]{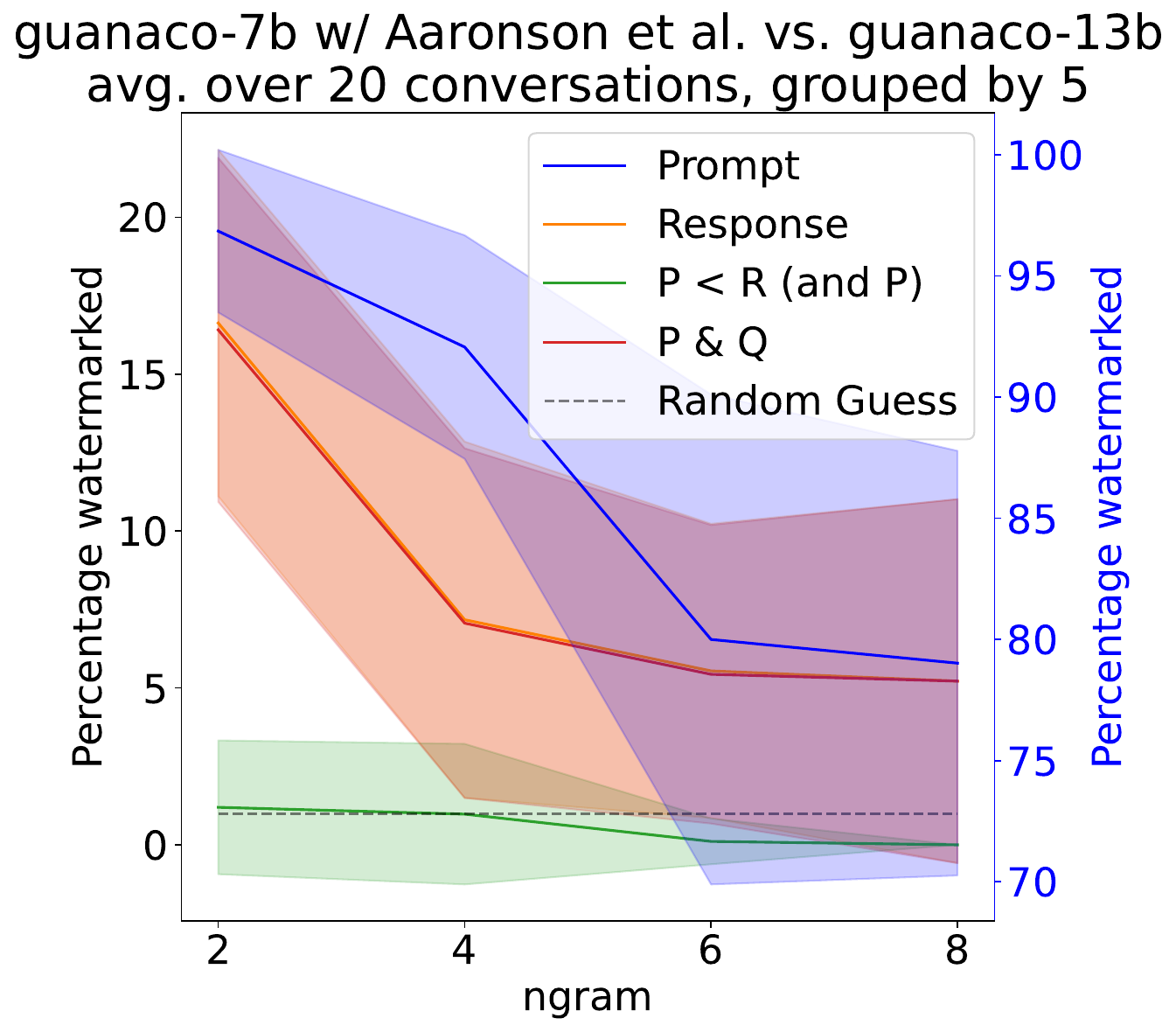}
\caption{\citet{aaronson2022my} with Guanco-7b (w) v 13b and varying ngrams, T=1}
\label{fig:aaronson_ngrams_7_13}
\end{figure*}

\begin{figure*}
\centering
\includegraphics[width=0.3\textwidth]{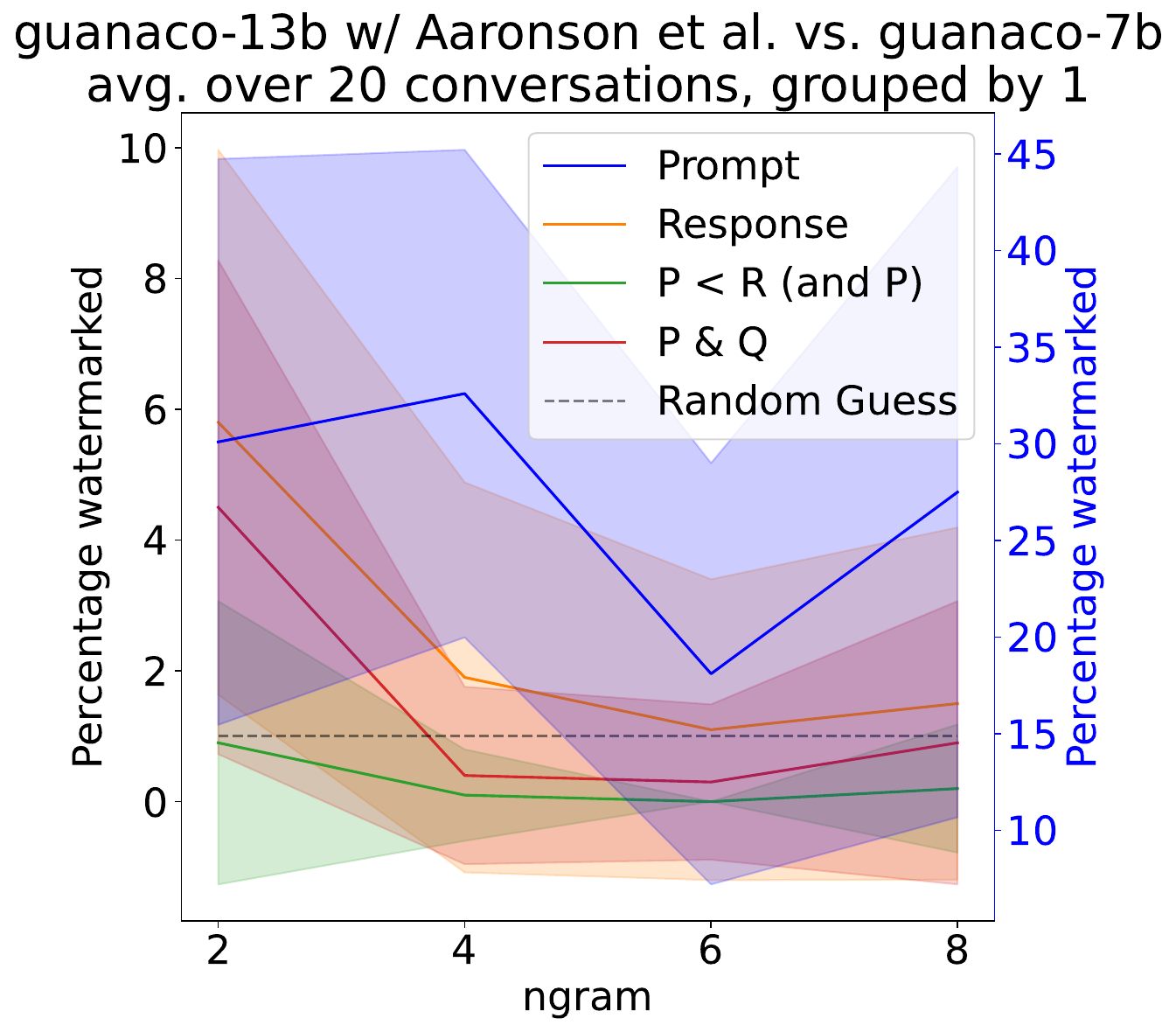}\hfill
\includegraphics[width=0.3\textwidth]{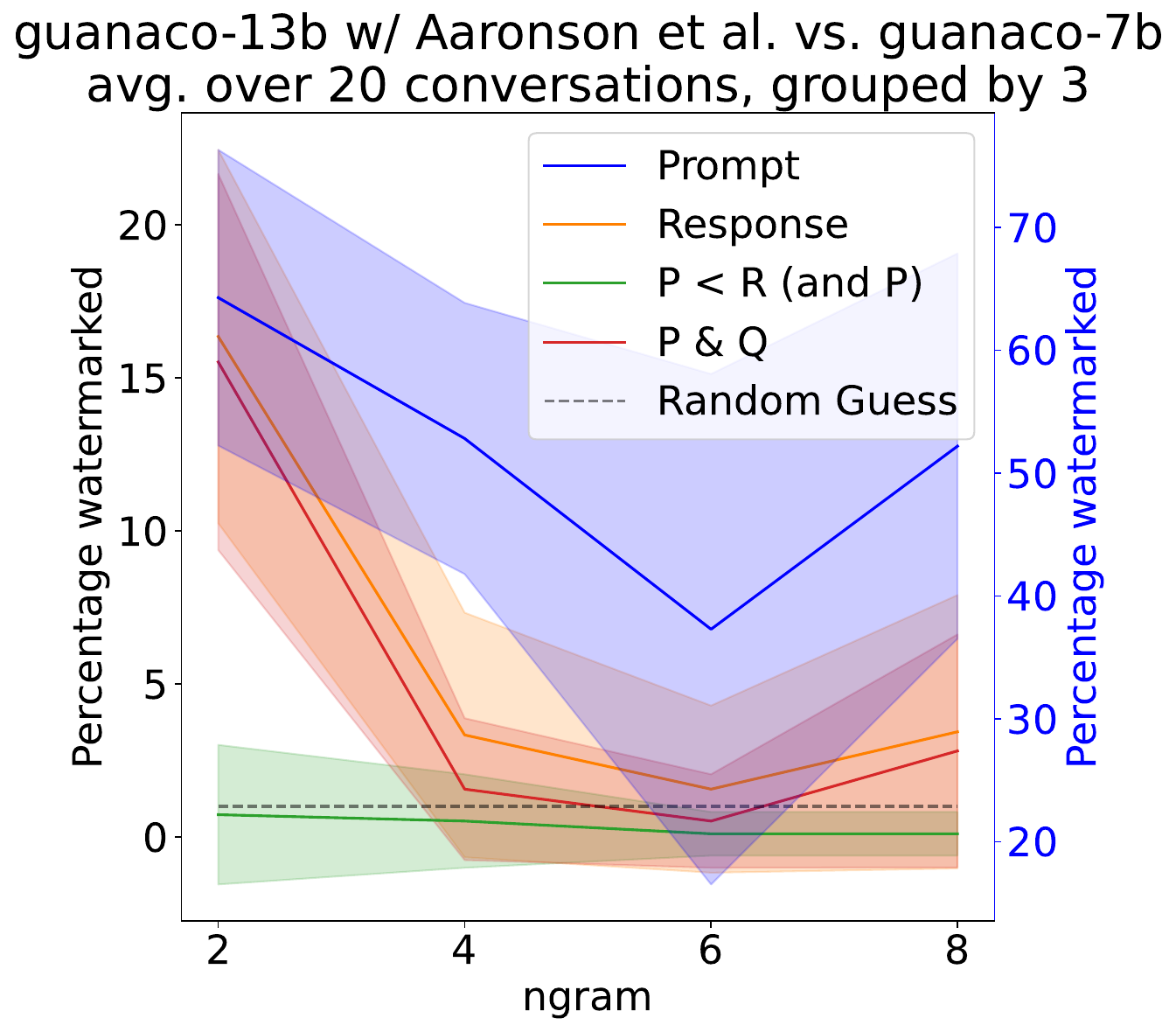}\hfill
\includegraphics[width=0.3\textwidth]{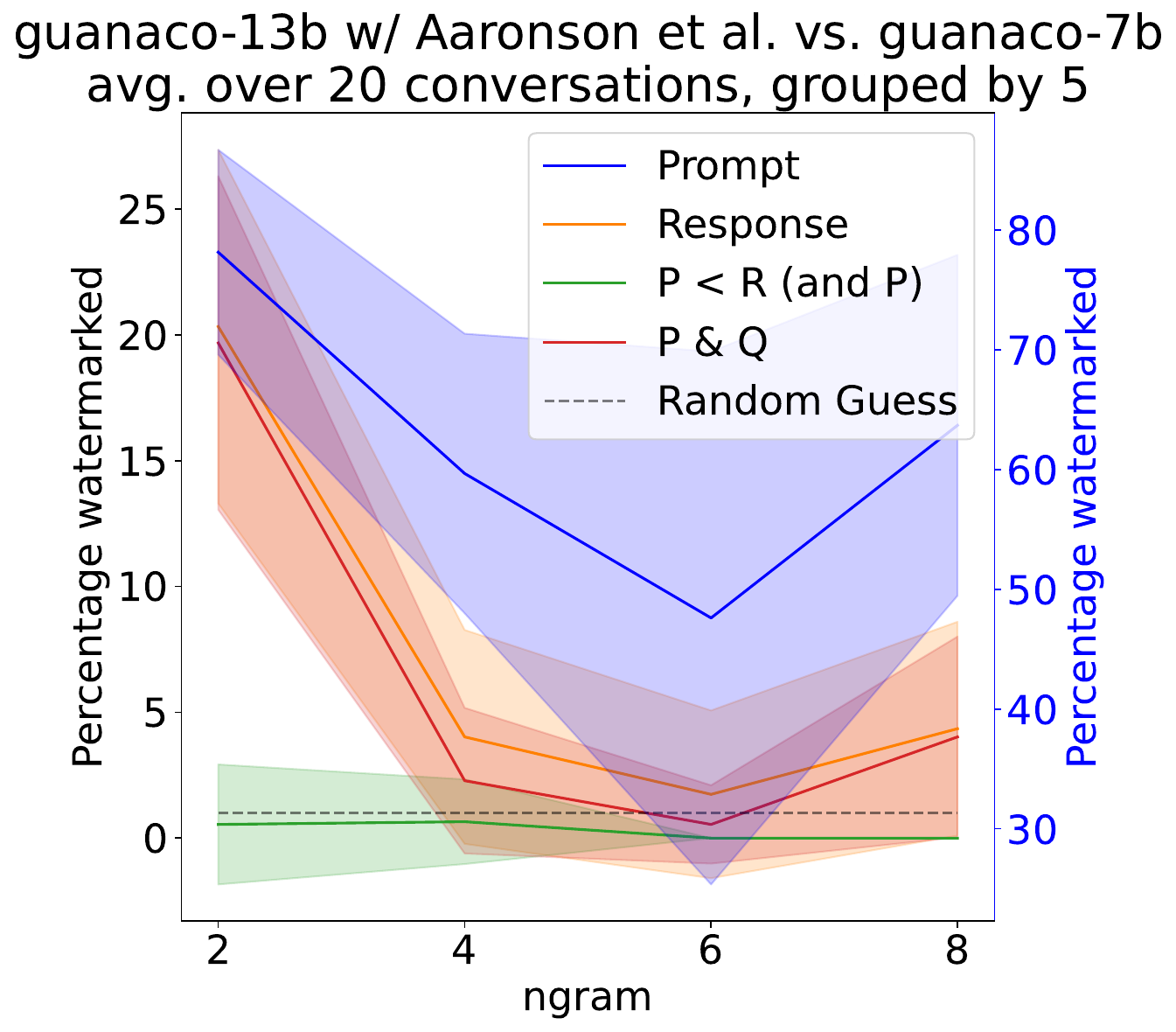}
\caption{\citet{aaronson2022my} with Guanco-13b (w) v 7b and varying ngrams, T=1}
\label{fig:aaronson_ngrams_13_7}
\end{figure*}

\begin{figure*}
\centering
\includegraphics[width=0.3\textwidth]{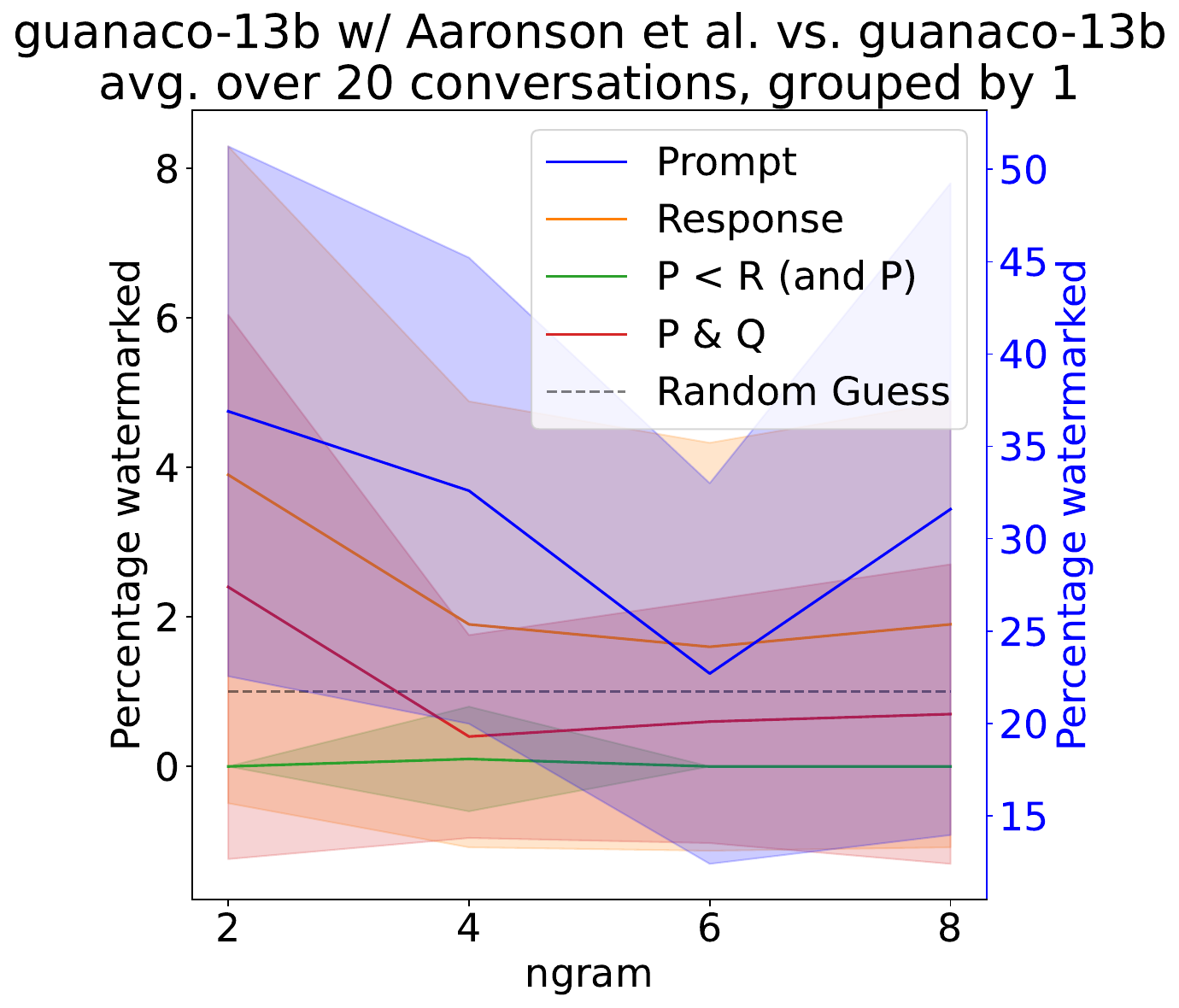}\hfill
\includegraphics[width=0.3\textwidth]{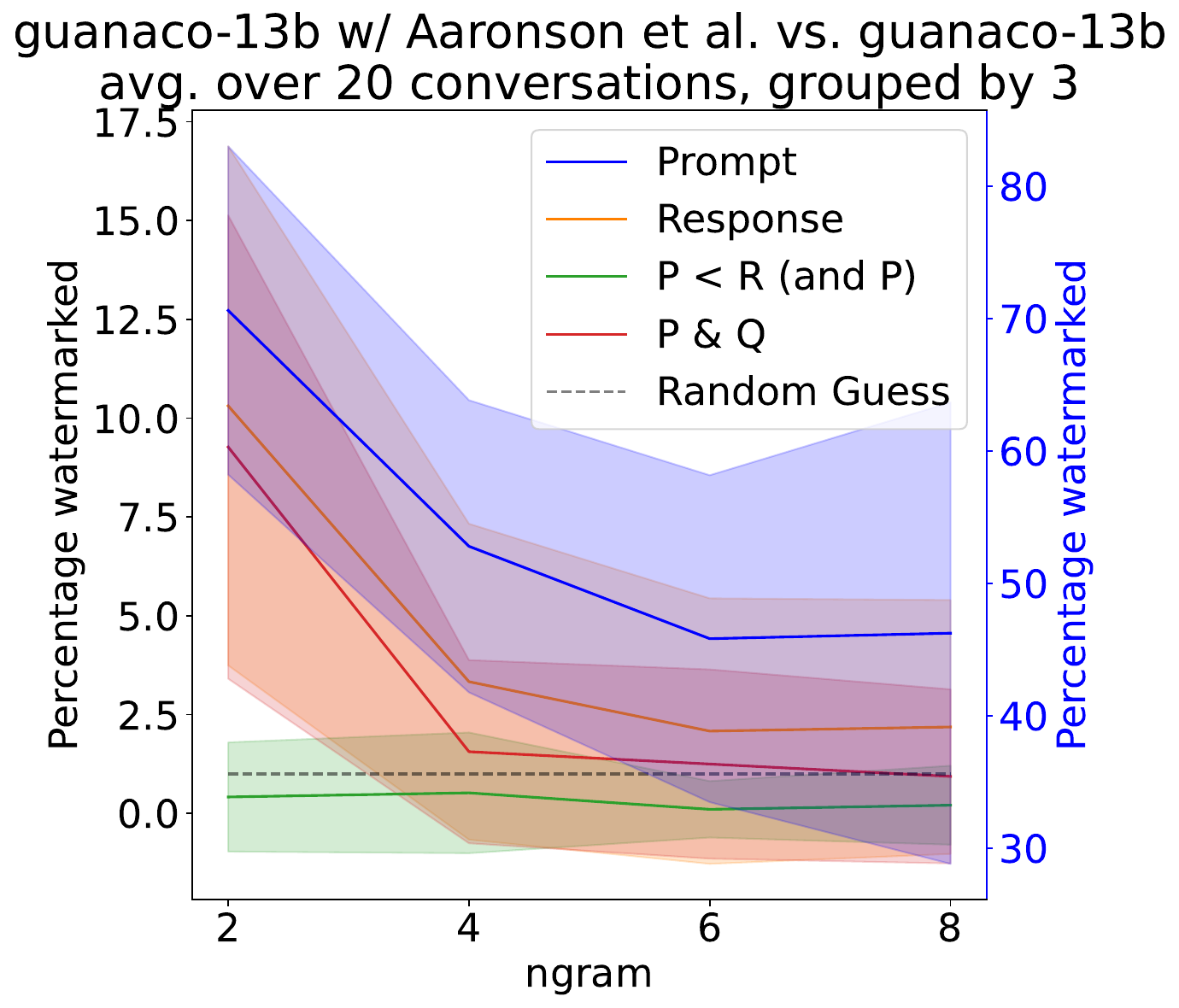}\hfill
\includegraphics[width=0.3\textwidth]{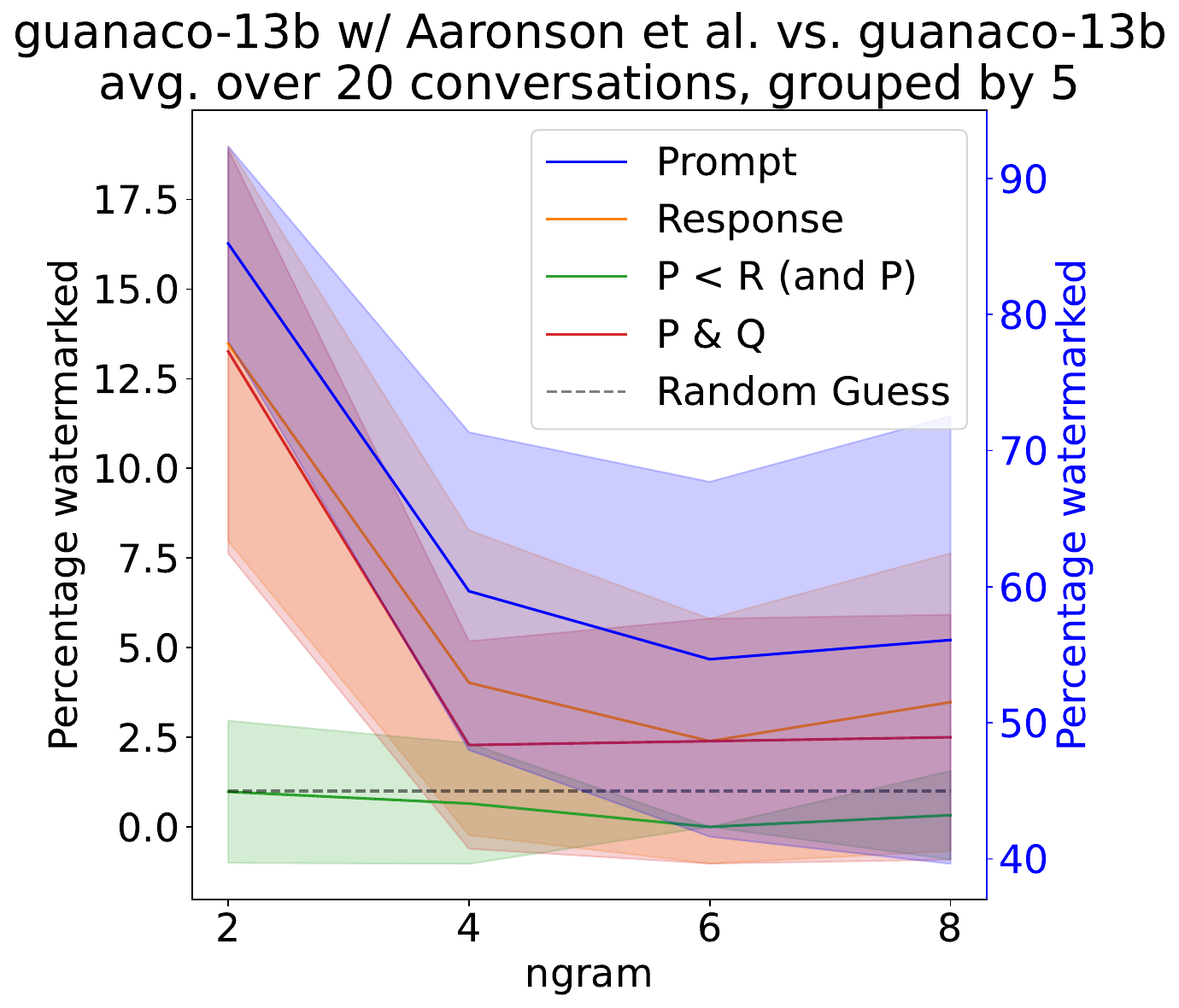}
\caption{\citet{aaronson2022my} watermarking with Guanco-13b (w) v 13b and varying ngrams, T=1}
\label{fig:aaronson_ngrams_13_13}
\end{figure*}

\begin{figure*}
\centering
\includegraphics[width=0.5\textwidth]{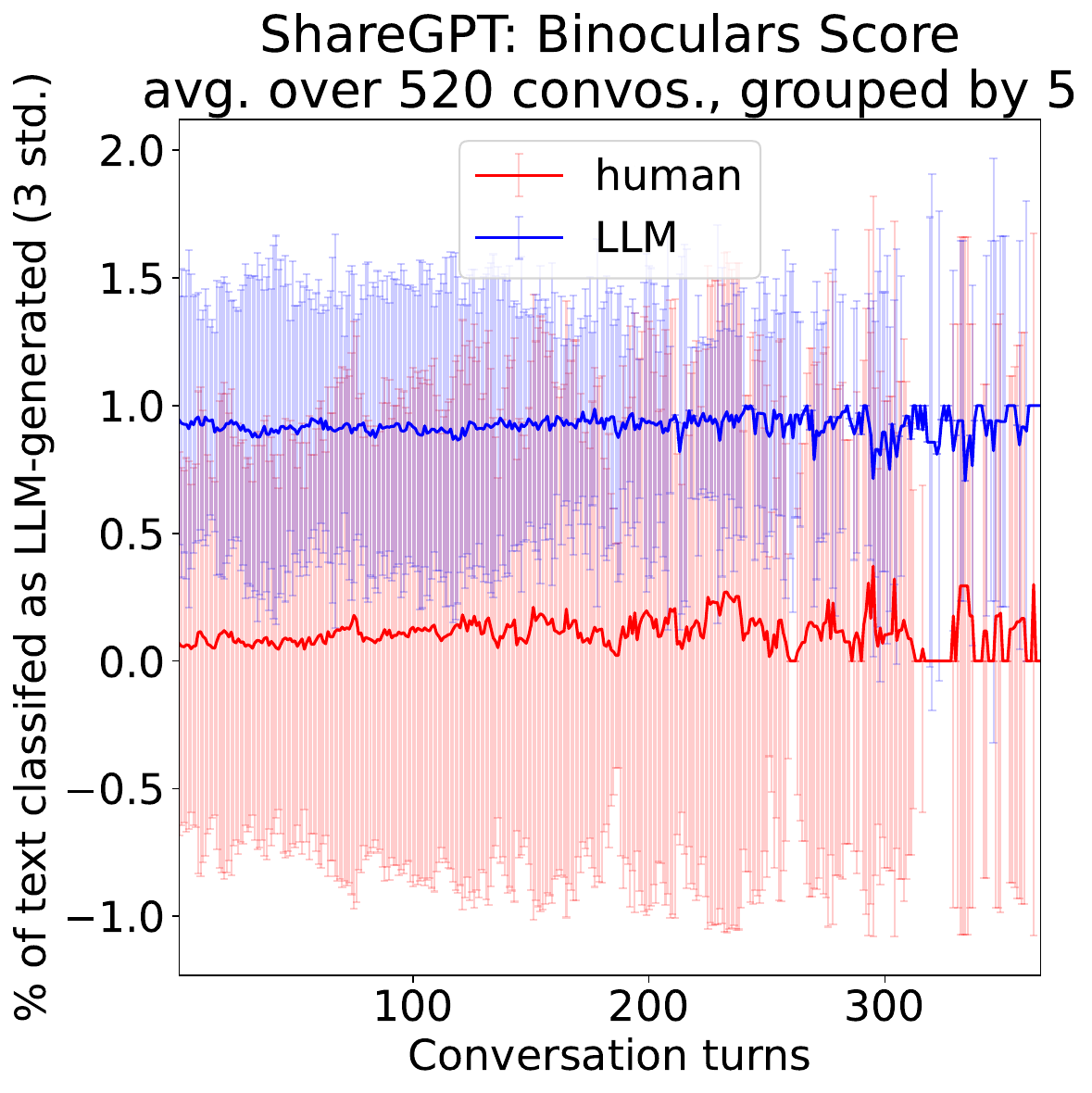}\hfill
\includegraphics[width=0.5\textwidth]{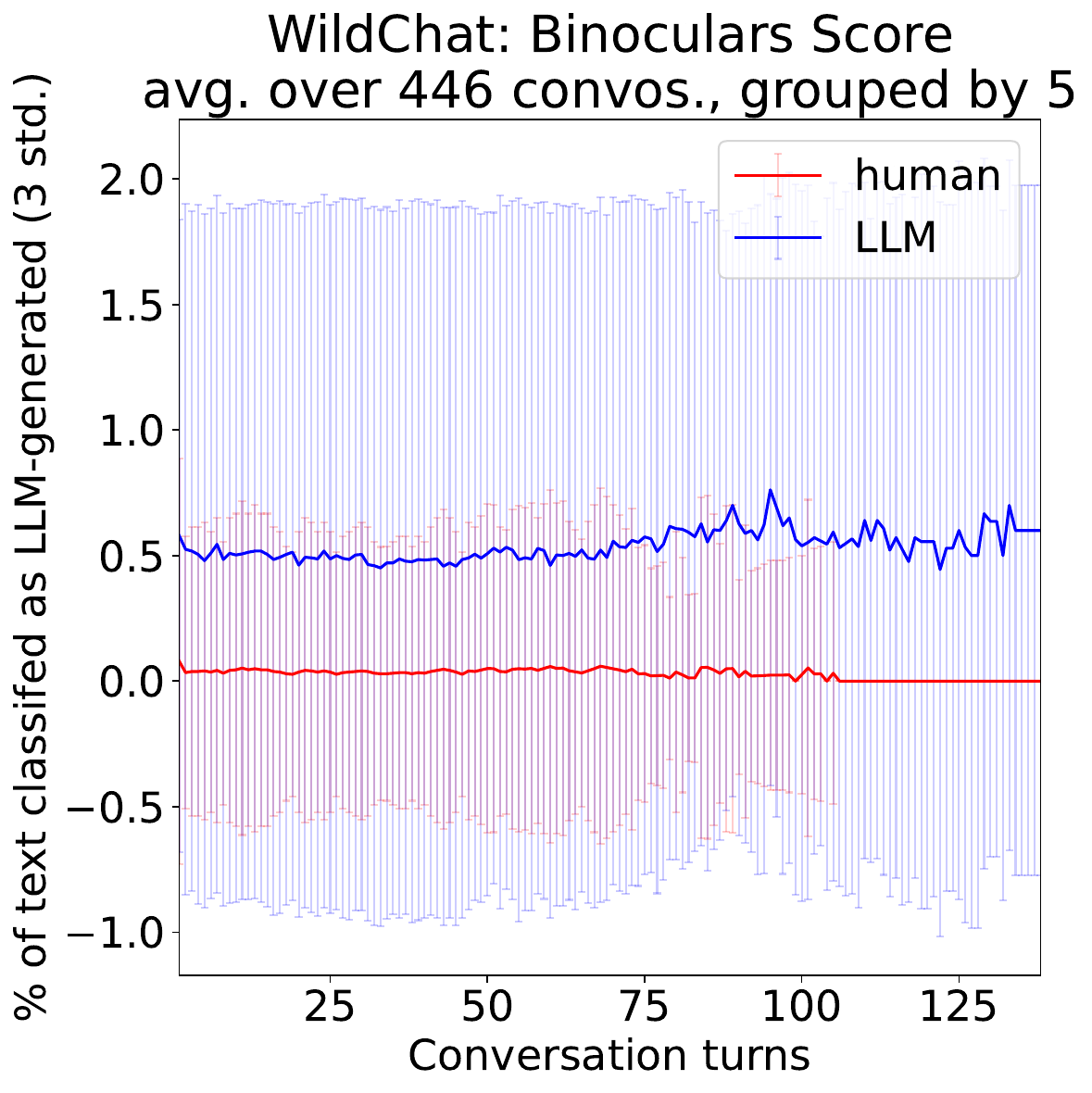}\hfill
\caption{Datasets split with Binoculars scoring. Showing three standard deviations. } 
\label{fig:sharegpt_wildchat}
\end{figure*}

\section{Original conversation prompt}
\label{sec:original_prompt}

\begin{lstlisting}
promptstart = "A chat between two people talking about everyday topics such as Weather, Family, Work/School, Hobbies/Interests, Entertainment, Travel, Food/Cooking, Current Events, Technology, Health/Fitness, Pets, Relationships, Home/Living, Fashion/Style, Goals/Ambitions.\nThe two people listen to each other intently and carry the conversation naturally and they talk at length. Answer with at least 50 words each time.\n\n### Person A: Hey there! How's the weather treating you lately? It's been so unpredictable here. One day it's sunny and warm, and the next it's pouring rain.\n\n### Person B: Oh, I know what you mean! The weather has been all over the place lately. It makes planning outdoor activities a bit tricky, doesn't it? I'm hoping for some more consistent sunshine soon so I can spend more time hiking and enjoying nature. How about you? Any outdoor plans in mind?"
\end{lstlisting}

\section{Watermarked conversation examples}
\label{sec:conversation_examples}

Below are a number of examples where both prompt (from watermarked model) and the response (from unwatermarked model) are watermarked. In colour we highlight the tokens that are reused from \textbf{P}rompt ({\color{ForestGreen}\mybox{green}}) into the \textbf{R}esponse ({\color{blue}\mybox{blue}}).

\noindent\textbf{P}: The holidays are {\color{ForestGreen}\mybox{a}} great time {\color{ForestGreen}\mybox{of}} year {\color{ForestGreen}\mybox{for}} reflection, {\color{ForestGreen}\mybox{and}} {\color{ForestGreen}\mybox{i}} {\color{ForestGreen}\mybox{was}} {\color{ForestGreen}\mybox{thinking}} {\color{ForestGreen}\mybox{about}} {\color{ForestGreen}\mybox{my}} {\color{ForestGreen}\mybox{new}} Year's {\color{ForestGreen}\mybox{resolutions}} {\color{ForestGreen}\mybox{for}} {\color{ForestGreen}\mybox{next}} year. Sounds like {\color{ForestGreen}\mybox{a}} plan! I'm {\color{ForestGreen}\mybox{not}} {\color{ForestGreen}\mybox{a}} {\color{ForestGreen}\mybox{fan}} {\color{ForestGreen}\mybox{of}} resolutions, either, {\color{ForestGreen}\mybox{but}} I'm {\color{ForestGreen}\mybox{up}} {\color{ForestGreen}\mybox{for}} {\color{ForestGreen}\mybox{a}} challenge. {\color{ForestGreen}\mybox{hiking}} {\color{ForestGreen}\mybox{a}} {\color{ForestGreen}\mybox{new}} {\color{ForestGreen}\mybox{trail}} {\color{ForestGreen}\mybox{is}} {\color{ForestGreen}\mybox{always}} {\color{ForestGreen}\mybox{a}} blast.
\\\textbf{R}: That's {\color{blue}\mybox{a}} good idea! I'm also {\color{blue}\mybox{thinking}} {\color{blue}\mybox{about}} {\color{blue}\mybox{my}} {\color{blue}\mybox{resolutions}} {\color{blue}\mybox{for}} {\color{blue}\mybox{next}} year. I'm {\color{blue}\mybox{not}} {\color{blue}\mybox{a}} {\color{blue}\mybox{fan}} {\color{blue}\mybox{of}} making them, {\color{blue}\mybox{but}} I'm {\color{blue}\mybox{always}} {\color{blue}\mybox{up}} {\color{blue}\mybox{for}} {\color{blue}\mybox{a}} challenge. I'm {\color{blue}\mybox{not}} {\color{blue}\mybox{a}} {\color{blue}\mybox{fan}} {\color{blue}\mybox{of}} making {\color{blue}\mybox{new}} Year's resolutions, {\color{blue}\mybox{but}} I'm {\color{blue}\mybox{always}} {\color{blue}\mybox{up}} {\color{blue}\mybox{for}} {\color{blue}\mybox{a}} challenge. {\color{blue}\mybox{i}} {\color{blue}\mybox{was}} {\color{blue}\mybox{thinking}} {\color{blue}\mybox{about}} trying something {\color{blue}\mybox{new}} this year, {\color{blue}\mybox{and}} {\color{blue}\mybox{i}} {\color{blue}\mybox{was}} wondering if you had any ideas. {\color{blue}\mybox{hiking}} {\color{blue}\mybox{is}} one {\color{blue}\mybox{of}} {\color{blue}\mybox{my}} favorite things to do, {\color{blue}\mybox{but}} I'm {\color{blue}\mybox{not}} {\color{blue}\mybox{a}} {\color{blue}\mybox{fan}} {\color{blue}\mybox{of}} doing {\color{blue}\mybox{the}} same {\color{blue}\mybox{trail}} over {\color{blue}\mybox{and}} over again.

\noindent\textbf{P}: Sounds like {\color{ForestGreen}\mybox{a}} plan! I'm {\color{ForestGreen}\mybox{not}} {\color{ForestGreen}\mybox{a}} {\color{ForestGreen}\mybox{fan}} {\color{ForestGreen}\mybox{of}} resolutions, either, {\color{ForestGreen}\mybox{but}} I'm {\color{ForestGreen}\mybox{up}} {\color{ForestGreen}\mybox{for}} {\color{ForestGreen}\mybox{a}} challenge. {\color{ForestGreen}\mybox{hiking}} {\color{ForestGreen}\mybox{a}} {\color{ForestGreen}\mybox{new}} {\color{ForestGreen}\mybox{trail}} {\color{ForestGreen}\mybox{is}} {\color{ForestGreen}\mybox{always}} {\color{ForestGreen}\mybox{a}} blast. Sounds like {\color{ForestGreen}\mybox{a}} plan! I'm {\color{ForestGreen}\mybox{not}} {\color{ForestGreen}\mybox{a}} {\color{ForestGreen}\mybox{fan}} {\color{ForestGreen}\mybox{of}} repetition, either, {\color{ForestGreen}\mybox{and}} I'm {\color{ForestGreen}\mybox{always}} {\color{ForestGreen}\mybox{up}} {\color{ForestGreen}\mybox{for}} {\color{ForestGreen}\mybox{a}} {\color{ForestGreen}\mybox{new}} adventure.
\\\textbf{R}: I'm {\color{blue}\mybox{not}} {\color{blue}\mybox{a}} {\color{blue}\mybox{fan}} {\color{blue}\mybox{of}} making {\color{blue}\mybox{new}} Year's resolutions, {\color{blue}\mybox{but}} I'm {\color{blue}\mybox{always}} {\color{blue}\mybox{up}} {\color{blue}\mybox{for}} {\color{blue}\mybox{a}} challenge. I was thinking about trying something {\color{blue}\mybox{new}} this year, {\color{blue}\mybox{and}} I was wondering if you had any ideas. {\color{blue}\mybox{hiking}} {\color{blue}\mybox{is}} one {\color{blue}\mybox{of}} my favorite things to do, {\color{blue}\mybox{but}} I'm {\color{blue}\mybox{not}} {\color{blue}\mybox{a}} {\color{blue}\mybox{fan}} {\color{blue}\mybox{of}} doing the same {\color{blue}\mybox{trail}} over {\color{blue}\mybox{and}} over again. I'm {\color{blue}\mybox{not}} {\color{blue}\mybox{a}} {\color{blue}\mybox{fan}} {\color{blue}\mybox{of}} repetition, either, {\color{blue}\mybox{and}} I'm {\color{blue}\mybox{always}} {\color{blue}\mybox{up}} {\color{blue}\mybox{for}} {\color{blue}\mybox{a}} {\color{blue}\mybox{new}} adventure. I was thinking about trying something {\color{blue}\mybox{new}} this year, {\color{blue}\mybox{and}} I was wondering if you had any ideas.

\noindent\textbf{P}: The weather is finally warming {\color{ForestGreen}\mybox{up}} here {\color{ForestGreen}\mybox{in}} [place]. I'm {\color{ForestGreen}\mybox{so}} {\color{ForestGreen}\mybox{excited}} {\color{ForestGreen}\mybox{to}} be {\color{ForestGreen}\mybox{able}} {\color{ForestGreen}\mybox{to}} go outside {\color{ForestGreen}\mybox{and}} enjoy {\color{ForestGreen}\mybox{the}} {\color{ForestGreen}\mybox{nice}} weather. I'm {\color{ForestGreen}\mybox{the}} {\color{ForestGreen}\mybox{same}} way. {\color{ForestGreen}\mybox{i}} {\color{ForestGreen}\mybox{love}} {\color{ForestGreen}\mybox{to}} {\color{ForestGreen}\mybox{keep}} {\color{ForestGreen}\mybox{up}} {\color{ForestGreen}\mybox{with}} {\color{ForestGreen}\mybox{the}} {\color{ForestGreen}\mybox{latest}} {\color{ForestGreen}\mybox{news}} {\color{ForestGreen}\mybox{and}} {\color{ForestGreen}\mybox{developments}} {\color{ForestGreen}\mybox{in}} [the field]. It's {\color{ForestGreen}\mybox{always}} {\color{ForestGreen}\mybox{nice}} {\color{ForestGreen}\mybox{to}} {\color{ForestGreen}\mybox{meet}} {\color{ForestGreen}\mybox{someone}} {\color{ForestGreen}\mybox{who}} {\color{ForestGreen}\mybox{shares}} {\color{ForestGreen}\mybox{your}} {\color{ForestGreen}\mybox{passions}} {\color{ForestGreen}\mybox{and}} interests. I'm {\color{ForestGreen}\mybox{so}} {\color{ForestGreen}\mybox{glad}} we're {\color{ForestGreen}\mybox{able}} {\color{ForestGreen}\mybox{to}} {\color{ForestGreen}\mybox{connect}} {\color{ForestGreen}\mybox{over}} {\color{ForestGreen}\mybox{our}} {\color{ForestGreen}\mybox{shared}} {\color{ForestGreen}\mybox{interest}} {\color{ForestGreen}\mybox{in}} [the field]. {\color{ForestGreen}\mybox{the}} election is right around {\color{ForestGreen}\mybox{the}} corner, {\color{ForestGreen}\mybox{and}} I'm {\color{ForestGreen}\mybox{so}} {\color{ForestGreen}\mybox{excited}} {\color{ForestGreen}\mybox{to}} see what happens. I'm {\color{ForestGreen}\mybox{so}} {\color{ForestGreen}\mybox{glad}} we're {\color{ForestGreen}\mybox{able}} {\color{ForestGreen}\mybox{to}} {\color{ForestGreen}\mybox{connect}} {\color{ForestGreen}\mybox{over}} {\color{ForestGreen}\mybox{our}} {\color{ForestGreen}\mybox{shared}} {\color{ForestGreen}\mybox{interest}} {\color{ForestGreen}\mybox{in}} politics.
\\\textbf{R}: I'm {\color{blue}\mybox{so}} {\color{blue}\mybox{excited}} for [the new thing]. I'm {\color{blue}\mybox{so}} {\color{blue}\mybox{glad}} we're {\color{blue}\mybox{able}} {\color{blue}\mybox{to}} {\color{blue}\mybox{connect}} {\color{blue}\mybox{over}} {\color{blue}\mybox{our}} {\color{blue}\mybox{shared}} interests. Me too. I'm {\color{blue}\mybox{so}} happy {\color{blue}\mybox{to}} have found {\color{blue}\mybox{someone}} {\color{blue}\mybox{who}} understands my {\color{blue}\mybox{love}} of [the field]. I'm {\color{blue}\mybox{the}} {\color{blue}\mybox{same}} way. {\color{blue}\mybox{i}} {\color{blue}\mybox{love}} {\color{blue}\mybox{to}} {\color{blue}\mybox{keep}} {\color{blue}\mybox{up}} {\color{blue}\mybox{with}} {\color{blue}\mybox{the}} {\color{blue}\mybox{latest}} {\color{blue}\mybox{news}} {\color{blue}\mybox{and}} {\color{blue}\mybox{developments}} {\color{blue}\mybox{in}} politics. It's {\color{blue}\mybox{always}} {\color{blue}\mybox{nice}} {\color{blue}\mybox{to}} {\color{blue}\mybox{meet}} {\color{blue}\mybox{someone}} {\color{blue}\mybox{who}} {\color{blue}\mybox{shares}} {\color{blue}\mybox{your}} {\color{blue}\mybox{passions}} {\color{blue}\mybox{and}} interests. I'm {\color{blue}\mybox{so}} {\color{blue}\mybox{glad}} we're {\color{blue}\mybox{able}} {\color{blue}\mybox{to}} {\color{blue}\mybox{connect}} {\color{blue}\mybox{over}} {\color{blue}\mybox{our}} {\color{blue}\mybox{shared}} {\color{blue}\mybox{interest}} {\color{blue}\mybox{in}} politics.

\noindent\textbf{P}: I'm {\color{ForestGreen}\mybox{the}} {\color{ForestGreen}\mybox{same}} way. {\color{ForestGreen}\mybox{i}} {\color{ForestGreen}\mybox{love}} {\color{ForestGreen}\mybox{to}} {\color{ForestGreen}\mybox{keep}} {\color{ForestGreen}\mybox{up}} {\color{ForestGreen}\mybox{with}} {\color{ForestGreen}\mybox{the}} {\color{ForestGreen}\mybox{latest}} {\color{ForestGreen}\mybox{news}} {\color{ForestGreen}\mybox{and}} {\color{ForestGreen}\mybox{developments}} {\color{ForestGreen}\mybox{in}} [the field]. It's {\color{ForestGreen}\mybox{always}} {\color{ForestGreen}\mybox{nice}} {\color{ForestGreen}\mybox{to}} {\color{ForestGreen}\mybox{meet}} {\color{ForestGreen}\mybox{someone}} {\color{ForestGreen}\mybox{who}} {\color{ForestGreen}\mybox{shares}} {\color{ForestGreen}\mybox{your}} {\color{ForestGreen}\mybox{passions}} {\color{ForestGreen}\mybox{and}} interests. I'm {\color{ForestGreen}\mybox{so}} {\color{ForestGreen}\mybox{glad}} we're {\color{ForestGreen}\mybox{able}} {\color{ForestGreen}\mybox{to}} {\color{ForestGreen}\mybox{connect}} {\color{ForestGreen}\mybox{over}} {\color{ForestGreen}\mybox{our}} {\color{ForestGreen}\mybox{shared}} {\color{ForestGreen}\mybox{interest}} {\color{ForestGreen}\mybox{in}} [the field]. {\color{ForestGreen}\mybox{the}} election is right around {\color{ForestGreen}\mybox{the}} corner, {\color{ForestGreen}\mybox{and}} I'm {\color{ForestGreen}\mybox{so}} {\color{ForestGreen}\mybox{excited}} {\color{ForestGreen}\mybox{to}} see what happens. I'm {\color{ForestGreen}\mybox{so}} {\color{ForestGreen}\mybox{glad}} we're {\color{ForestGreen}\mybox{able}} {\color{ForestGreen}\mybox{to}} {\color{ForestGreen}\mybox{connect}} {\color{ForestGreen}\mybox{over}} {\color{ForestGreen}\mybox{our}} {\color{ForestGreen}\mybox{shared}} {\color{ForestGreen}\mybox{interest}} {\color{ForestGreen}\mybox{in}} politics. {\color{ForestGreen}\mybox{the}} weather is finally warming {\color{ForestGreen}\mybox{up}} here {\color{ForestGreen}\mybox{in}} [place]. I'm {\color{ForestGreen}\mybox{so}} {\color{ForestGreen}\mybox{excited}} {\color{ForestGreen}\mybox{to}} be {\color{ForestGreen}\mybox{able}} {\color{ForestGreen}\mybox{to}} go outside {\color{ForestGreen}\mybox{and}} enjoy {\color{ForestGreen}\mybox{the}} {\color{ForestGreen}\mybox{nice}} weather.
\\\textbf{R}: Me too. I'm {\color{blue}\mybox{so}} happy {\color{blue}\mybox{to}} have found {\color{blue}\mybox{someone}} {\color{blue}\mybox{who}} understands my {\color{blue}\mybox{love}} of [the field]. I'm {\color{blue}\mybox{the}} {\color{blue}\mybox{same}} way. {\color{blue}\mybox{i}} {\color{blue}\mybox{love}} {\color{blue}\mybox{to}} {\color{blue}\mybox{keep}} {\color{blue}\mybox{up}} {\color{blue}\mybox{with}} {\color{blue}\mybox{the}} {\color{blue}\mybox{latest}} {\color{blue}\mybox{news}} {\color{blue}\mybox{and}} {\color{blue}\mybox{developments}} {\color{blue}\mybox{in}} politics. It's {\color{blue}\mybox{always}} {\color{blue}\mybox{nice}} {\color{blue}\mybox{to}} {\color{blue}\mybox{meet}} {\color{blue}\mybox{someone}} {\color{blue}\mybox{who}} {\color{blue}\mybox{shares}} {\color{blue}\mybox{your}} {\color{blue}\mybox{passions}} {\color{blue}\mybox{and}} interests. I'm {\color{blue}\mybox{so}} {\color{blue}\mybox{glad}} we're {\color{blue}\mybox{able}} {\color{blue}\mybox{to}} {\color{blue}\mybox{connect}} {\color{blue}\mybox{over}} {\color{blue}\mybox{our}} {\color{blue}\mybox{shared}} {\color{blue}\mybox{interest}} {\color{blue}\mybox{in}} politics. I'm {\color{blue}\mybox{so}} {\color{blue}\mybox{excited}} for [the new thing]. I'm {\color{blue}\mybox{so}} {\color{blue}\mybox{glad}} we're {\color{blue}\mybox{able}} {\color{blue}\mybox{to}} {\color{blue}\mybox{connect}} {\color{blue}\mybox{over}} {\color{blue}\mybox{our}} {\color{blue}\mybox{shared}} interests.

\noindent\textbf{P}: The weather is finally warming up here in [place]. I'm {\color{ForestGreen}\mybox{so}} {\color{ForestGreen}\mybox{excited}} {\color{ForestGreen}\mybox{to}} be {\color{ForestGreen}\mybox{able}} {\color{ForestGreen}\mybox{to}} go outside and enjoy the nice weather. The election is right around the corner, and I'm {\color{ForestGreen}\mybox{so}} {\color{ForestGreen}\mybox{excited}} {\color{ForestGreen}\mybox{to}} see what happens. I'm {\color{ForestGreen}\mybox{so}} {\color{ForestGreen}\mybox{glad}} we're {\color{ForestGreen}\mybox{able}} {\color{ForestGreen}\mybox{to}} {\color{ForestGreen}\mybox{connect}} {\color{ForestGreen}\mybox{over}} {\color{ForestGreen}\mybox{our}} {\color{ForestGreen}\mybox{shared}} interest in politics. [Topic] is {\color{ForestGreen}\mybox{so}} fascinating {\color{ForestGreen}\mybox{to}} me as well. I'm {\color{ForestGreen}\mybox{so}} {\color{ForestGreen}\mybox{glad}} we're {\color{ForestGreen}\mybox{able}} {\color{ForestGreen}\mybox{to}} {\color{ForestGreen}\mybox{connect}} {\color{ForestGreen}\mybox{over}} {\color{ForestGreen}\mybox{our}} {\color{ForestGreen}\mybox{shared}} interest in [topic].
\\\textbf{R}: I'm {\color{blue}\mybox{so}} {\color{blue}\mybox{excited}} for [the new thing]. I'm {\color{blue}\mybox{so}} {\color{blue}\mybox{glad}} we're {\color{blue}\mybox{able}} {\color{blue}\mybox{to}} {\color{blue}\mybox{connect}} {\color{blue}\mybox{over}} {\color{blue}\mybox{our}} {\color{blue}\mybox{shared}} interests. I'm also really into [topic]. I'm {\color{blue}\mybox{so}} {\color{blue}\mybox{glad}} we're {\color{blue}\mybox{able}} {\color{blue}\mybox{to}} {\color{blue}\mybox{connect}} {\color{blue}\mybox{over}} {\color{blue}\mybox{our}} {\color{blue}\mybox{shared}} interests. I'm {\color{blue}\mybox{so}} {\color{blue}\mybox{excited}} about [new thing]. I'm {\color{blue}\mybox{so}} {\color{blue}\mybox{glad}} we're {\color{blue}\mybox{able}} {\color{blue}\mybox{to}} {\color{blue}\mybox{connect}} {\color{blue}\mybox{over}} {\color{blue}\mybox{our}} {\color{blue}\mybox{shared}} interests.

\noindent\textbf{P}: The election {\color{ForestGreen}\mybox{is}} right around the corner, and I'm {\color{ForestGreen}\mybox{so}} {\color{ForestGreen}\mybox{excited}} {\color{ForestGreen}\mybox{to}} see what happens. I'm {\color{ForestGreen}\mybox{so}} {\color{ForestGreen}\mybox{glad}} we're {\color{ForestGreen}\mybox{able}} {\color{ForestGreen}\mybox{to}} {\color{ForestGreen}\mybox{connect}} {\color{ForestGreen}\mybox{over}} {\color{ForestGreen}\mybox{our}} {\color{ForestGreen}\mybox{shared}} {\color{ForestGreen}\mybox{interest}} {\color{ForestGreen}\mybox{in}} politics. [Topic] {\color{ForestGreen}\mybox{is}} {\color{ForestGreen}\mybox{so}} {\color{ForestGreen}\mybox{fascinating}} {\color{ForestGreen}\mybox{to}} {\color{ForestGreen}\mybox{me}} {\color{ForestGreen}\mybox{as}} well. I'm {\color{ForestGreen}\mybox{so}} {\color{ForestGreen}\mybox{glad}} we're {\color{ForestGreen}\mybox{able}} {\color{ForestGreen}\mybox{to}} {\color{ForestGreen}\mybox{connect}} {\color{ForestGreen}\mybox{over}} {\color{ForestGreen}\mybox{our}} {\color{ForestGreen}\mybox{shared}} {\color{ForestGreen}\mybox{interest}} {\color{ForestGreen}\mybox{in}} [topic]. The weather {\color{ForestGreen}\mybox{is}} finally warming up here {\color{ForestGreen}\mybox{in}} [place]. I'm {\color{ForestGreen}\mybox{so}} {\color{ForestGreen}\mybox{excited}} {\color{ForestGreen}\mybox{to}} be {\color{ForestGreen}\mybox{able}} {\color{ForestGreen}\mybox{to}} go outside and enjoy the nice weather.
\\\textbf{R}: I'm also really into [topic]. I'm {\color{blue}\mybox{so}} {\color{blue}\mybox{glad}} we're {\color{blue}\mybox{able}} {\color{blue}\mybox{to}} {\color{blue}\mybox{connect}} {\color{blue}\mybox{over}} {\color{blue}\mybox{our}} {\color{blue}\mybox{shared}} interests. I'm {\color{blue}\mybox{so}} {\color{blue}\mybox{excited}} about [new thing]. I'm {\color{blue}\mybox{so}} {\color{blue}\mybox{glad}} we're {\color{blue}\mybox{able}} {\color{blue}\mybox{to}} {\color{blue}\mybox{connect}} {\color{blue}\mybox{over}} {\color{blue}\mybox{our}} {\color{blue}\mybox{shared}} interests. [new topic] {\color{blue}\mybox{is}} {\color{blue}\mybox{so}} {\color{blue}\mybox{fascinating}} {\color{blue}\mybox{to}} {\color{blue}\mybox{me}} {\color{blue}\mybox{as}} well. I'm {\color{blue}\mybox{so}} {\color{blue}\mybox{glad}} we're {\color{blue}\mybox{able}} {\color{blue}\mybox{to}} {\color{blue}\mybox{connect}} {\color{blue}\mybox{over}} {\color{blue}\mybox{our}} {\color{blue}\mybox{shared}} {\color{blue}\mybox{interest}} {\color{blue}\mybox{in}} [topic].

\noindent\textbf{P}: The most important thing in {\color{ForestGreen}\mybox{my}} life {\color{ForestGreen}\mybox{is}} {\color{ForestGreen}\mybox{my}} family {\color{ForestGreen}\mybox{and}} {\color{ForestGreen}\mybox{the}} friends {\color{ForestGreen}\mybox{i}} consider {\color{ForestGreen}\mybox{my}} family.  {\color{ForestGreen}\mybox{the}} most important thing in {\color{ForestGreen}\mybox{my}} life {\color{ForestGreen}\mybox{is}} {\color{ForestGreen}\mybox{my}} family {\color{ForestGreen}\mybox{and}} {\color{ForestGreen}\mybox{the}} friends {\color{ForestGreen}\mybox{i}} consider {\color{ForestGreen}\mybox{my}} family.  {\color{ForestGreen}\mybox{the}} most important thing in {\color{ForestGreen}\mybox{my}} life {\color{ForestGreen}\mybox{is}} {\color{ForestGreen}\mybox{my}} family {\color{ForestGreen}\mybox{and}} {\color{ForestGreen}\mybox{the}} friends {\color{ForestGreen}\mybox{i}} consider {\color{ForestGreen}\mybox{my}} family. 
\\\textbf{R}: I'm a big fan of soccer, especially {\color{blue}\mybox{the}} Premier League. I'm a supporter of Manchester United. {\color{blue}\mybox{my}} favorite player {\color{blue}\mybox{is}} Paul Pogba. I'm a big fan of anime. {\color{blue}\mybox{my}} favorite anime {\color{blue}\mybox{is}} Fullmetal Alchemist: Brotherhood. {\color{blue}\mybox{i}} want to travel {\color{blue}\mybox{the}} world. {\color{blue}\mybox{i}} want to see as many places as {\color{blue}\mybox{i}} can {\color{blue}\mybox{and}} experience different cultures.

\noindent\textbf{P}: The most important thing in {\color{ForestGreen}\mybox{my}} life {\color{ForestGreen}\mybox{is}} {\color{ForestGreen}\mybox{my}} family {\color{ForestGreen}\mybox{and}} {\color{ForestGreen}\mybox{the}} friends {\color{ForestGreen}\mybox{i}} consider {\color{ForestGreen}\mybox{my}} family.  {\color{ForestGreen}\mybox{the}} most important thing in {\color{ForestGreen}\mybox{my}} life {\color{ForestGreen}\mybox{is}} {\color{ForestGreen}\mybox{my}} family {\color{ForestGreen}\mybox{and}} {\color{ForestGreen}\mybox{the}} friends {\color{ForestGreen}\mybox{i}} consider {\color{ForestGreen}\mybox{my}} family.  {\color{ForestGreen}\mybox{the}} most important thing in {\color{ForestGreen}\mybox{my}} life {\color{ForestGreen}\mybox{is}} {\color{ForestGreen}\mybox{my}} family {\color{ForestGreen}\mybox{and}} {\color{ForestGreen}\mybox{the}} friends {\color{ForestGreen}\mybox{i}} consider {\color{ForestGreen}\mybox{my}} family. 
\\\textbf{R}: I'm a big fan of anime. {\color{blue}\mybox{my}} favorite anime {\color{blue}\mybox{is}} Fullmetal Alchemist: Brotherhood. {\color{blue}\mybox{i}} want to travel {\color{blue}\mybox{the}} world. {\color{blue}\mybox{i}} want to see as many places as {\color{blue}\mybox{i}} can {\color{blue}\mybox{and}} experience different cultures. I'm a sucker for good-looking cars. {\color{blue}\mybox{my}} dream car {\color{blue}\mybox{is}} a Lamborghini Huracán.

\noindent\textbf{P}: (New topic) {\color{ForestGreen}\mybox{is}} {\color{ForestGreen}\mybox{a}} {\color{ForestGreen}\mybox{very}} {\color{ForestGreen}\mybox{important}} (field) {\color{ForestGreen}\mybox{in}} today's world. The weather {\color{ForestGreen}\mybox{is}} beautiful these days, don't you think? (New topic) {\color{ForestGreen}\mybox{is}} {\color{ForestGreen}\mybox{a}} {\color{ForestGreen}\mybox{very}} {\color{ForestGreen}\mybox{important}} (field) {\color{ForestGreen}\mybox{in}} today's world.
\\\textbf{R}: (New topic) has {\color{blue}\mybox{a}} huge impact on our society. I've been learning {\color{blue}\mybox{a}} lot about (new topic). It's so interesting! (New topic) {\color{blue}\mybox{is}} {\color{blue}\mybox{a}} {\color{blue}\mybox{very}} {\color{blue}\mybox{important}} (field) {\color{blue}\mybox{in}} today's world.

\noindent\textbf{P}: You know, {\color{ForestGreen}\mybox{as}} {\color{ForestGreen}\mybox{much}} {\color{ForestGreen}\mybox{as}} {\color{ForestGreen}\mybox{i}} {\color{ForestGreen}\mybox{love}} {\color{ForestGreen}\mybox{being}} outdoors, {\color{ForestGreen}\mybox{i}} {\color{ForestGreen}\mybox{also}} {\color{ForestGreen}\mybox{love}} {\color{ForestGreen}\mybox{taking}} {\color{ForestGreen}\mybox{time}} {\color{ForestGreen}\mybox{to}} {\color{ForestGreen}\mybox{just}} {\color{ForestGreen}\mybox{relax}} {\color{ForestGreen}\mybox{and}} unwind. I'm {\color{ForestGreen}\mybox{always}} {\color{ForestGreen}\mybox{telling}} {\color{ForestGreen}\mybox{my}} {\color{ForestGreen}\mybox{friends}} {\color{ForestGreen}\mybox{how}} {\color{ForestGreen}\mybox{much}} {\color{ForestGreen}\mybox{i}} {\color{ForestGreen}\mybox{love}} {\color{ForestGreen}\mybox{a}} {\color{ForestGreen}\mybox{nice}} {\color{ForestGreen}\mybox{glass}} {\color{ForestGreen}\mybox{of}} {\color{ForestGreen}\mybox{wine}} {\color{ForestGreen}\mybox{and}} {\color{ForestGreen}\mybox{a}} {\color{ForestGreen}\mybox{good}} book. I'm {\color{ForestGreen}\mybox{the}} same way. {\color{ForestGreen}\mybox{i}} {\color{ForestGreen}\mybox{love}} {\color{ForestGreen}\mybox{the}} holidays {\color{ForestGreen}\mybox{and}} I'm {\color{ForestGreen}\mybox{always}} {\color{ForestGreen}\mybox{so}} {\color{ForestGreen}\mybox{excited}} {\color{ForestGreen}\mybox{to}} see {\color{ForestGreen}\mybox{my}} {\color{ForestGreen}\mybox{family}} {\color{ForestGreen}\mybox{and}} friends. I'm {\color{ForestGreen}\mybox{the}} same way. {\color{ForestGreen}\mybox{i}} {\color{ForestGreen}\mybox{love}} {\color{ForestGreen}\mybox{to}} bake {\color{ForestGreen}\mybox{and}} I'm {\color{ForestGreen}\mybox{always}} {\color{ForestGreen}\mybox{looking}} {\color{ForestGreen}\mybox{forward}} {\color{ForestGreen}\mybox{to}} {\color{ForestGreen}\mybox{decorating}} {\color{ForestGreen}\mybox{the}} {\color{ForestGreen}\mybox{house}} {\color{ForestGreen}\mybox{for}} {\color{ForestGreen}\mybox{the}} holidays.
\\\textbf{R}: I'm {\color{blue}\mybox{so}} {\color{blue}\mybox{excited}} {\color{blue}\mybox{for}} {\color{blue}\mybox{the}} holidays! {\color{blue}\mybox{i}} {\color{blue}\mybox{love}} this {\color{blue}\mybox{time}} {\color{blue}\mybox{of}} year {\color{blue}\mybox{and}} I'm {\color{blue}\mybox{so}} {\color{blue}\mybox{looking}} {\color{blue}\mybox{forward}} {\color{blue}\mybox{to}} spending {\color{blue}\mybox{time}} with {\color{blue}\mybox{family}} {\color{blue}\mybox{and}} friends. Me too! I'm {\color{blue}\mybox{looking}} {\color{blue}\mybox{forward}} {\color{blue}\mybox{to}} baking cookies {\color{blue}\mybox{and}} {\color{blue}\mybox{decorating}} {\color{blue}\mybox{the}} {\color{blue}\mybox{house}} {\color{blue}\mybox{for}} {\color{blue}\mybox{the}} holidays. {\color{blue}\mybox{you}} know, {\color{blue}\mybox{as}} {\color{blue}\mybox{much}} {\color{blue}\mybox{as}} {\color{blue}\mybox{i}} {\color{blue}\mybox{love}} {\color{blue}\mybox{being}} outdoors, {\color{blue}\mybox{i}} {\color{blue}\mybox{also}} {\color{blue}\mybox{love}} {\color{blue}\mybox{taking}} {\color{blue}\mybox{time}} {\color{blue}\mybox{to}} {\color{blue}\mybox{just}} {\color{blue}\mybox{relax}} {\color{blue}\mybox{and}} unwind. I'm {\color{blue}\mybox{always}} {\color{blue}\mybox{telling}} {\color{blue}\mybox{my}} {\color{blue}\mybox{friends}} {\color{blue}\mybox{how}} {\color{blue}\mybox{much}} {\color{blue}\mybox{i}} {\color{blue}\mybox{love}} {\color{blue}\mybox{a}} {\color{blue}\mybox{nice}} {\color{blue}\mybox{glass}} {\color{blue}\mybox{of}} {\color{blue}\mybox{wine}} {\color{blue}\mybox{and}} {\color{blue}\mybox{a}} {\color{blue}\mybox{good}} book.

\noindent\textbf{P}: I'm {\color{ForestGreen}\mybox{the}} same way. {\color{ForestGreen}\mybox{i}} {\color{ForestGreen}\mybox{love}} {\color{ForestGreen}\mybox{to}} bake {\color{ForestGreen}\mybox{and}} I'm {\color{ForestGreen}\mybox{always}} {\color{ForestGreen}\mybox{looking}} {\color{ForestGreen}\mybox{forward}} {\color{ForestGreen}\mybox{to}} decorating {\color{ForestGreen}\mybox{the}} house {\color{ForestGreen}\mybox{for}} {\color{ForestGreen}\mybox{the}} holidays. You know, {\color{ForestGreen}\mybox{as}} {\color{ForestGreen}\mybox{much}} {\color{ForestGreen}\mybox{as}} {\color{ForestGreen}\mybox{i}} {\color{ForestGreen}\mybox{love}} {\color{ForestGreen}\mybox{being}} outdoors, {\color{ForestGreen}\mybox{i}} {\color{ForestGreen}\mybox{also}} {\color{ForestGreen}\mybox{love}} {\color{ForestGreen}\mybox{taking}} {\color{ForestGreen}\mybox{time}} {\color{ForestGreen}\mybox{to}} {\color{ForestGreen}\mybox{just}} {\color{ForestGreen}\mybox{relax}} {\color{ForestGreen}\mybox{and}} unwind. I'm {\color{ForestGreen}\mybox{always}} {\color{ForestGreen}\mybox{telling}} {\color{ForestGreen}\mybox{my}} {\color{ForestGreen}\mybox{friends}} {\color{ForestGreen}\mybox{how}} {\color{ForestGreen}\mybox{much}} {\color{ForestGreen}\mybox{i}} {\color{ForestGreen}\mybox{love}} {\color{ForestGreen}\mybox{a}} {\color{ForestGreen}\mybox{nice}} {\color{ForestGreen}\mybox{glass}} {\color{ForestGreen}\mybox{of}} {\color{ForestGreen}\mybox{wine}} {\color{ForestGreen}\mybox{and}} {\color{ForestGreen}\mybox{a}} {\color{ForestGreen}\mybox{good}} book. {\color{ForestGreen}\mybox{the}} weather is beautiful today! {\color{ForestGreen}\mybox{i}} was {\color{ForestGreen}\mybox{just}} thinking about all {\color{ForestGreen}\mybox{the}} fun summer activities {\color{ForestGreen}\mybox{i}} can't wait {\color{ForestGreen}\mybox{to}} do.
\\\textbf{R}: You know, {\color{blue}\mybox{as}} {\color{blue}\mybox{much}} {\color{blue}\mybox{as}} {\color{blue}\mybox{i}} {\color{blue}\mybox{love}} {\color{blue}\mybox{being}} outdoors, {\color{blue}\mybox{i}} {\color{blue}\mybox{also}} {\color{blue}\mybox{love}} {\color{blue}\mybox{taking}} {\color{blue}\mybox{time}} {\color{blue}\mybox{to}} {\color{blue}\mybox{just}} {\color{blue}\mybox{relax}} {\color{blue}\mybox{and}} unwind. I'm {\color{blue}\mybox{always}} {\color{blue}\mybox{telling}} {\color{blue}\mybox{my}} {\color{blue}\mybox{friends}} {\color{blue}\mybox{how}} {\color{blue}\mybox{much}} {\color{blue}\mybox{i}} {\color{blue}\mybox{love}} {\color{blue}\mybox{a}} {\color{blue}\mybox{nice}} {\color{blue}\mybox{glass}} {\color{blue}\mybox{of}} {\color{blue}\mybox{wine}} {\color{blue}\mybox{and}} {\color{blue}\mybox{a}} {\color{blue}\mybox{good}} book. I'm so excited {\color{blue}\mybox{for}} {\color{blue}\mybox{the}} holidays! {\color{blue}\mybox{i}} {\color{blue}\mybox{love}} this {\color{blue}\mybox{time}} {\color{blue}\mybox{of}} year {\color{blue}\mybox{and}} I'm so {\color{blue}\mybox{looking}} {\color{blue}\mybox{forward}} {\color{blue}\mybox{to}} spending {\color{blue}\mybox{time}} with family {\color{blue}\mybox{and}} friends. {\color{blue}\mybox{i}} know, right? It's like they're so happy {\color{blue}\mybox{to}} see us {\color{blue}\mybox{and}} they never tire {\color{blue}\mybox{of}} us {\color{blue}\mybox{being}} home.

\noindent\textbf{P}: You know, as much as {\color{ForestGreen}\mybox{i}} {\color{ForestGreen}\mybox{love}} {\color{ForestGreen}\mybox{being}} outdoors, {\color{ForestGreen}\mybox{i}} also {\color{ForestGreen}\mybox{love}} taking {\color{ForestGreen}\mybox{time}} {\color{ForestGreen}\mybox{to}} just relax {\color{ForestGreen}\mybox{and}} unwind. I'm always telling my friends how much {\color{ForestGreen}\mybox{i}} {\color{ForestGreen}\mybox{love}} a nice glass {\color{ForestGreen}\mybox{of}} wine {\color{ForestGreen}\mybox{and}} a good book. {\color{ForestGreen}\mybox{the}} weather is beautiful today! {\color{ForestGreen}\mybox{i}} was just thinking about all {\color{ForestGreen}\mybox{the}} fun summer activities {\color{ForestGreen}\mybox{i}} can't wait {\color{ForestGreen}\mybox{to}} do. {\color{ForestGreen}\mybox{the}} weather is beautiful today! {\color{ForestGreen}\mybox{i}} was just thinking about all {\color{ForestGreen}\mybox{the}} fun summer activities {\color{ForestGreen}\mybox{i}} can't wait {\color{ForestGreen}\mybox{to}} do.
\\\textbf{R}: I'm so excited for {\color{blue}\mybox{the}} holidays! {\color{blue}\mybox{i}} {\color{blue}\mybox{love}} this {\color{blue}\mybox{time}} {\color{blue}\mybox{of}} year {\color{blue}\mybox{and}} I'm so looking forward {\color{blue}\mybox{to}} spending {\color{blue}\mybox{time}} with family {\color{blue}\mybox{and}} friends. {\color{blue}\mybox{i}} know, right? It's like they're so happy {\color{blue}\mybox{to}} see us {\color{blue}\mybox{and}} they never tire {\color{blue}\mybox{of}} us {\color{blue}\mybox{being}} home. {\color{blue}\mybox{i}} know, right? It's like they're so happy {\color{blue}\mybox{to}} see us {\color{blue}\mybox{and}} they never tire {\color{blue}\mybox{of}} us {\color{blue}\mybox{being}} home.

\noindent\textbf{P}: Sounds like {\color{ForestGreen}\mybox{a}} great idea. I'll {\color{ForestGreen}\mybox{get}} {\color{ForestGreen}\mybox{you}} {\color{ForestGreen}\mybox{a}} {\color{ForestGreen}\mybox{new}} car. I'm {\color{ForestGreen}\mybox{so}} glad you're adopting {\color{ForestGreen}\mybox{a}} puppy. [topics: car, babysitter, gift] {\color{ForestGreen}\mybox{the}} babysitter told me they had {\color{ForestGreen}\mybox{a}} bumper car accident {\color{ForestGreen}\mybox{with}} our {\color{ForestGreen}\mybox{new}} car.
\\\textbf{R}: We're {\color{blue}\mybox{so}} excited! [topics: car, babysitter, gift] {\color{blue}\mybox{the}} tutor we hired was great. She really helped my son {\color{blue}\mybox{with}} his history. I'm {\color{blue}\mybox{so}} sorry to hear that. I hope no one was hurt. I'll {\color{blue}\mybox{get}} {\color{blue}\mybox{you}} {\color{blue}\mybox{a}} {\color{blue}\mybox{new}} car.

\noindent\textbf{P}: I'm {\color{ForestGreen}\mybox{so}} glad you're adopting {\color{ForestGreen}\mybox{a}} puppy. [topics: car, babysitter, gift] The babysitter told me they had {\color{ForestGreen}\mybox{a}} bumper car accident {\color{ForestGreen}\mybox{with}} our {\color{ForestGreen}\mybox{new}} car. The babysitter told me they had {\color{ForestGreen}\mybox{a}} bumper car accident {\color{ForestGreen}\mybox{with}} our {\color{ForestGreen}\mybox{new}} car.
\\\textbf{R}: The tutor we hired was great. She really helped my son {\color{blue}\mybox{with}} his history. I'm {\color{blue}\mybox{so}} sorry to hear that. I hope no one was hurt. I'll get you {\color{blue}\mybox{a}} {\color{blue}\mybox{new}} car. My husband and I are thinking of moving to {\color{blue}\mybox{a}} {\color{blue}\mybox{new}} house.

\noindent\textbf{P}: The babysitter told me they had {\color{ForestGreen}\mybox{a}} bumper car accident with our {\color{ForestGreen}\mybox{new}} car. The babysitter told me they had {\color{ForestGreen}\mybox{a}} bumper car accident with our {\color{ForestGreen}\mybox{new}} car. The babysitter told me they had {\color{ForestGreen}\mybox{a}} bumper car accident with our {\color{ForestGreen}\mybox{new}} car.
\\\textbf{R}: I'm so sorry to hear that. I hope no one was hurt. I'll get you {\color{blue}\mybox{a}} {\color{blue}\mybox{new}} car. My husband and I are thinking of moving to {\color{blue}\mybox{a}} {\color{blue}\mybox{new}} house. My daughter is taking {\color{blue}\mybox{a}} dance class, and she's really enjoying it.

\section{Expanded table of results}
\label{sec:more_results}

\begin{table*}[h]
    \centering
    \adjustbox{max width=\linewidth}{
    \begin{tabular}{lllrrrrrrrr}
    \toprule
        \multicolumn{3}{l}{\textbf{Unwatermarked}} & 
        \multicolumn{4}{l}{\citet{aaronson2023watermarking}} & 
        \multicolumn{4}{r}{\citet{kirchenbauer2023awatermark}} \\ 
        \cmidrule(lr){4-7} \cmidrule(lr){8-11} 
        \textbf{Guanco} & \textbf{T} & \textbf{\#R} & 
        \textbf{P}rompt $\uparrow$ & \textbf{R}esponse $\downarrow$ & \textbf{P}\&\textbf{R} & \textbf{P}<\textbf{R} & 
        \textbf{P}rompt $\uparrow$ & \textbf{R}esponse $\downarrow$ & \textbf{P}\&\textbf{R} & \textbf{P}<\textbf{R} \\
    \midrule
Random & & & $1\%$ & $1\%$ & $0.01\%$ & & $1\%$ & $1\%$ & $0.01\%$ \\
\midrule
\multirow{11}{*}{7b (w) v 7b} 
& 0.0 & 1 & 0.6\% & 0.5\% & 0.0\% & 0.0\% & 68.5\% & 3.6\% & 3.2\% & 0.6\% \\
& 0.5 & 1 & 5.9\% & 1.6\% & 0.9\% & 0.1\% & 73.5\% & 6.8\% & 6.3\% & 0.5\% \\
& 1.0 & 1 & 49.9\% & 3.7\% & 2.9\% & 0.3\% & 82.4\% & 6.3\% & 6.0\% & 0.4\% \\ \\
& 0.0 & 3 & 1.0\% & 0.8\% & 0.1\% & 0.0\% & 91.1\% & 7.3\% & 7.1\% & 1.0\% \\
& 0.5 & 3 & 15.3\% & 3.3\% & 2.5\% & 0.8\% & 93.1\% & 10.2\% & 9.9\% & 1.0\% \\
& 1.0 & 3 & 72.0\% & 5.3\% & 4.9\% & 0.8\% & 96.7\% & 10.0\% & 10.0\% & 0.7\% \\ \\
& 0.0 & 5 & 1.4\% & 1.1\% & 0.2\% & 0.1\% & 95.4\% & 9.3\% & 9.1\% & 0.8\% \\
& 0.5 & 5 & 21.6\% & 4.0\% & 3.0\% & 0.9\% & 97.2\% & 13.4\% & 13.3\% & 1.2\% \\
& 1.0 & 5 & 81.1\% & 6.6\% & 6.4\% & 0.7\% & 98.9\% & 13.0\% & 13.0\% & 0.7\% \\
\midrule
\multirow{11}{*}{13b (w) v 13b} 
& 0.0 & 1 & 0.7\% & 0.7\% & 0.0\% & 0.0\% & 55.9\% & 3.1\% & 2.8\% & 0.8\% \\
& 0.5 & 1 & 3.4\% & 0.8\% & 0.1\% & 0.0\% & 57.4\% & 4.7\% & 3.8\% & 0.4\% \\
& 1.0 & 1 & 30.6\% & 3.2\% & 2.0\% & 0.3\% & 73.1\% & 5.8\% & 5.6\% & 0.4\% \\ \\
& 0.0 & 3 & 1.2\% & 1.8\% & 0.2\% & 0.1\% & 80.3\% & 5.6\% & 5.4\% & 0.7\% \\
& 0.5 & 3 & 10.3\% & 1.7\% & 0.8\% & 0.1\% & 79.8\% & 8.5\% & 7.8\% & 0.9\% \\
& 1.0 & 3 & 53.2\% & 6.3\% & 5.2\% & 0.7\% & 90.5\% & 8.3\% & 8.3\% & 0.4\% \\ \\
& 0.0 & 5 & 2.2\% & 3.0\% & 0.7\% & 0.3\% & 86.2\% & 7.6\% & 7.5\% & 0.7\% \\
& 0.5 & 5 & 17.8\% & 3.4\% & 2.1\% & 0.4\% & 86.3\% & 12.4\% & 11.7\% & 1.1\% \\
& 1.0 & 5 & 63.1\% & 8.1\% & 7.2\% & 0.9\% & 94.7\% & 11.2\% & 11.0\% & 0.4\% \\
\midrule

\multirow{11}{*}{7b (w) v 13b} 
& 0.0 & 1 & 0.4\% & 0.5\% & 0.1\% & 0.1\% & 67.5\% & 3.9\% & 3.2\% & 0.2\% \\
& 0.5 & 1 & 3.4\% & 1.2\% & 0.0\% & 0.0\% & 77.7\% & 2.3\% & 2.1\% & 0.2\% \\
& 1.0 & 1 & 63.4\% & 3.4\% & 2.9\% & 1.1\% & 83.8\% & 7.6\% & 7.4\% & 0.1\% \\ \\
& 0.0 & 3 & 2.6\% & 1.0\% & 0.1\% & 0.1\% & 89.6\% & 9.2\% & 8.8\% & 1.8\% \\
& 0.5 & 3 & 9.3\% & 2.4\% & 0.5\% & 0.0\% & 95.9\% & 5.5\% & 5.1\% & 0.0\% \\
& 1.0 & 3 & 80.0\% & 4.7\% & 4.5\% & 0.1\% & 97.7\% & 14.4\% & 14.1\% & 0.4\% \\ \\
& 0.0 & 5 & 4.1\% & 2.1\% & 0.7\% & 0.2\% & 96.0\% & 11.8\% & 11.8\% & 2.0\% \\
& 0.5 & 5 & 13.0\% & 3.3\% & 0.9\% & 0.1\% & 98.7\% & 8.0\% & 7.5\% & 0.0\% \\
& 1.0 & 5 & 89.3\% & 6.0\% & 6.0\% & 0.4\% & 99.3\% & 18.2\% & 18.2\% & 0.4\% \\
\midrule
\multirow{11}{*}{13b (w) v 7b} 
& 0.0 & 1 & 0.9\% & 0.8\% & 0.0\% & 0.0\% & 65.5\% & 4.1\% & 4.0\% & 0.4\% \\
& 0.5 & 1 & 3.1\% & 0.7\% & 0.1\% & 0.0\% & 66.9\% & 1.9\% & 1.7\% & 0.1\% \\
& 1.0 & 1 & 33.2\% & 3.9\% & 3.1\% & 0.4\% & 74.4\% & 4.6\% & 4.1\% & 0.4\% \\ \\
& 0.0 & 3 & 1.0\% & 0.1\% & 0.0\% & 0.0\% & 91.1\% & 6.8\% & 6.6\% & 0.3\% \\
& 0.5 & 3 & 11.0\% & 2.4\% & 0.9\% & 0.4\% & 92.1\% & 5.1\% & 4.7\% & 0.4\% \\
& 1.0 & 3 & 54.1\% & 6.7\% & 6.2\% & 1.1\% & 93.0\% & 7.4\% & 7.3\% & 0.7\% \\ \\
& 0.0 & 5 & 4.2\% & 0.7\% & 0.0\% & 0.0\% & 94.1\% & 7.9\% & 7.8\% & 0.1\% \\
& 0.5 & 5 & 15.3\% & 3.7\% & 1.7\% & 0.7\% & 96.1\% & 6.7\% & 6.3\% & 0.4\% \\
& 1.0 & 5 & 69.0\% & 8.4\% & 7.9\% & 0.5\% & 98.0\% & 10.2\% & 10.2\% & 0.8\% \\
    \bottomrule
    \end{tabular}}
    \caption{Expanded version of~\Cref{tab:combined_wms}. Table shows proportion of text that has a watermark. Here a watermarked LLM model (watermark ngram=4, threshold 0.01, averaged over three random keys (large variance)) interacts with a non-watermarked LLM model (Temperature=0.8). \textbf{P}rompt refers to the \textit{watermarked prompt} from a first model. \textbf{R}esponse refers to the \textit{response of an unwatermarked model} to a watermarked prompt. \textbf{\#R} refers to a number of responses concatenated together. \textbf{P\&R} reports a proportion of cases where watermarked prompt -- i.e. watermark score < 0.01 -- results in a watermarked response -- watermark score < 0.01. \textbf{P<R} reports the proportion of cases where watermark of the response from an unwatermarked model is stronger than the watermark in the prompt. Original conversation prompt is listed in~\Cref{sec:original_prompt}. }
    \label{tab:combined_wms_expanded}
\end{table*}

\end{document}